# Reconstruction of Simulation-Based Physical Field by Reconstruction Neural Network Method

Yu Li[a, *], Hu Wang[a, **], Kangjia Mo[a] and Tao Zeng[a]
[a] *State Key Laboratory of Advanced Design and Manufacturing for Vehicle Body, Hunan University, Changsha, 410082, PR China*

**Abstract.** A variety of modeling techniques have been developed in the past decade to reduce the computational expense and improve the accuracy of modeling. In this study, a new framework of modeling is suggested. Compared with other popular methods, a distinctive characteristic is "from image based model to analysis based model (e.g. stress, strain, and deformation)". In such a framework, a Reconstruction Neural Network (ReConNN) model designed for simulation-based physical field's reconstruction is proposed. The ReConNN contains two submodels that are Convolutional Neural Network (CNN) and Generative Adversarial Network (GAN). The CNN is employed to construct the mapping between contour images of physical field and objective function. Subsequently, the GAN is utilized to generate more images which are similar to the existing contour images. Finally, Lagrange polynomial is applied to complete the reconstruction. However, the existing CNN models are commonly applied to the classification tasks, which seem to be difficult to handle with regression tasks of images. Meanwhile, the existing GAN architectures are insufficient to generate high-accuracy "pseudo contour images". Therefore, a ReConNN model based on a Convolution in Convolution (CIC) and a Convolutional AutoEncoder-based Wasserstein Generative Adversarial Network (WGAN-CAE) is suggested. To evaluate the performance of the proposed model representatively, a classical topology optimization procedure is considered. Then the ReConNN is utilized to the reconstruction of heat transfer process of a pin fin heat sink. It demonstrates that the proposed ReConNN model is proved to be a potential capability to reconstruct physical field for multidisciplinary, such as structural optimization.

[*] Fist author: Y. Li. E-mail: liyu@hnu.edu.cn
[**] Corresponding author: H. Wang, State Key Laboratory of Advanced Design and Manufacturing for Vehicle Body, Hunan University, Changsha, 410082, PR China. E-mail: wanghu@hnu.edu.cn

## 1. Introduction

At present, many engineering analyses ask for the requirement of complicated and computationally expensive analysis and huge simulation codes (1), such as finite element analysis (FEA). A popular way to save the expensive computational cost is to generate an approximation of the complex analysis that describes the process accurately while at a much lower cost, which is usually called metamodel (2) and provides a "model of the model". Mathematically, assuming the input to the actual analysis is $x$, and the output is $y$, the true analysis code evaluates

$$y = f\left(x\right) \tag{1}$$

where $f(x)$ is a complex function. The metamodel approximation can be presented as

$$\hat{y} = g\left(x\right) \tag{2}$$

such that

$$y = \hat{y} + \varepsilon \tag{3}$$

where $\varepsilon$ includes both approximation and random errors.

There are multiple kinds of metamodeling techniques to approximate $f(x)$ using $g(x)$, e.g. Polynomial Regression (PR) (3, 4), Multivariate Adaptive Regression Spline (MARS) (5), Radial Basis Function (RBF) (6), Kriging (KG) (2, 7), and Support Vector Regression (SVR) (1, 8).

Different from the mentioned techniques based on "from analysis based model to objective function" where the analysis based model includes stress, strain, deformation and so on, this study proposed another framework "from image based model to analysis based model" which uses images derived from simulation results to reconstruct the physical field. In this study, a neural network model, which is named ReConNN, based on the framework to reconstruct the physical field is proposed. The ReConNN contains two submodels, one is CNN and another is GAN.

In the ReConNN, the CNN which is employed to construct the mapping between contour images and objective function is a kind of supervised learning neural networks. The CNN is a well-known deep learning architecture inspired by the natural visual perception mechanism of the living creatures. In particular, the CNN is able to achieve state-of-the-art results in classification tasks (9) and has been successfully utilized to many different tasks including speech processing (10, 11), image recognition (12-14), object detection (15-17), and text recognition (18, 19). However, when the CNN is applied to regression tasks such as constructing the mapping between contour images and objective function, the existing CNN models might be powerless (20).

Therefore, to improve accuracy without increasing computational cost when the task is regression, a CIC submodel is proposed where the image segmentation has been employed in order to improve accuracy through increasing network parameters. Subsequently, in order to reconstruct the physical field high-accurately, GAN is considered.

The GAN (21) is a powerful class of generative models that cast generative modeling as a game between two networks: the generator $G(z)$ which maps a sample $z$ depending on a random or a Gaussian distribution to the data distribution, and the discriminator $D(x)$ which determines if a sample $x$ belongs to the data distribution. While, the existing GAN models face many unsolved difficulties such as being difficult trained. In this study, according to experiments, it is found to be hard to generate satisfied images using the existing GAN models.

Therefore, a WGAN-CAE submodel is proposed. Convolutional AutoEncoder (CAE) (22) is a unsupervised learning and consisted of encoder $E(y)$ which compresses input samples to a smaller size, and decoder $D(x)$ which restores the compressed samples to their original size. The CAE equals the input sample $y$ and output value $\hat{y}$ through back propagation (BP) algorithm. Mathematically, the CAE can be described as

$$E\left(y\right) = x \tag{4}$$

$$D\left(x\right) = \hat{y} \tag{5}$$

$$y = \hat{y} + \varepsilon \tag{6}$$

where $x$ is intermediate variable between the encoder and decoder which has a much smaller size than $y$, and $\varepsilon$ includes both approximation and random errors.

In the WGAN-CAE, learning objects of WGAN are changed to the compressed samples, and the generated values will be restored using the CAE's decoder.

The remainder of this paper is organized as follows. In Section 2, some closely related works are reviewed. Next, the proposed ReConNN model are presented in Section 3. The detailed experimental methodology, results, and analysis are shown in Section 4. After a series of observations and analyses, the ReConNN is applied to the reconstruction of heat transfer process of a pin fin heat sink in Section 5. The summaries are given in the final section.

## 2. Related works

### *2.1. Existing metamedoling techniques*

There currently exist a number of metamodeling techniques in the last 20 years.

Response Surface Methodology (RSM) (23) approximates functions by using the least squares method on a series of points in the design variable space. Low-order polynomials are the most widely used response surface approximating functions, and first-order and second-order polynomial are calculated by Eqs. (7) and (8) respectively.

$$\hat{y} = b_0 + \sum_{i=1}^{k} b_i x_i \tag{7}$$

$$\hat{y} = b_0 + \sum_{i=1}^{k} b_i x_i + \sum_{i=1}^{k} b_{ii} x_i^2 + \sum\sum_{i<j} b_{ij} x_i x_j \tag{8}$$

where the constants ($b_0$, $b_i$, $b_{ii}$, $b_{ij}$) are determined by least squares regression.

Radial Basis Function (RBF) (23) attempts approximation by using a linear combination of radially symmetric functions. The RBF has produced good approximations to arbitrary contours. Mathematically, the model can be expressed as

$$\hat{y} = \sum_{i} a_i \left\| \mathbf{X} - \mathbf{X}_{0i} \right\| \tag{9}$$

where $a_i$ is a real-valued weight, and $\mathbf{X}_{0i}$ is the input vector.

Kriging (KG) (24, 25) postulates a combination of a known function and departures of the form.

$$y(x) = f(x) + Z(x) \tag{10}$$

where $f(x)$ is a polynomial function which is often taken as a constant, and $Z(x)$ is the correlation function which is a realization of a stochastic process with mean zero, variance $\sigma^2$, and nonzero covariance.

Moving Least Squares Method (MLSM) (26) is a model building technique that has been suggested for the use in the meshless form of the finite element method. It is a generalization of a traditional weighted least squares model building where weights do not remain constants but are functions of Euclidian distance from a sampling point to a point $x$ where the surrogate model is evaluated.

Multivariate Adaptive Regression Splines (MARS) (5) is a nonparametric regression procedure that makes no assumption about the underlying functional relationship between dependent and independent variables. Instead, the MARS constructs this relation from a set of coefficients and basis functions that are determined from regression data. The input space is divided into regions containing their own regression equation. Thus, the MARS is suitable for problems with high input dimensions, where the curse of dimensionality would likely create problems for other techniques.

Support Vector Machine (SVM) (27) is a binary linear classification technique in machine learning, which separates the classes with largest gap (called optimal margin) between the border line instances (called Support Vectors). The SVM has been extended to multi-class problems and has been extended using Kernels for non-linearly separable data problems.

### *2.2. Existing CNN models*

Inspired by Hubel and Wiesel's breakthrough findings in cells of animal visual cortex (28, 29), Fukushima (30) proposed a hierarchical model called Neocognitron which could be regarded as the predecessor of CNN. The first CNN architecture is proposed by LeCun (31, 32) in 1990. However, due to lack of large training samples and short of computing power at that time, the CNN couldn't perform well on very complex problems. Until 2012, with the appearance of AlexNet (33), the advantages of local connection, weight sharing, and local pooling of CNNs are widely recognized. With the success of AlexNet, many models have been proposed to improve its performance, e.g. ZFNet (34), VGGNet (35), GoogleNet (36), ResNet (37), InceptionNet (38, 39), Network in Network (40), and so on. Since then, the study of the CNN can be mainly divided into four directions: optimization of network architecture (35, 41, 42), enhancement of convolutional layer (34, 38-40), more attention of detection task (43-45), and add of new architectures (46-48).

From the evolution of architectures, a typical trend is that the networks are getting deeper. However, by increasing deeper, it increases the complexity of the network, which makes the network be more difficult to optimize and easier to be overfitting (9, 17, 49). Along this way, various methods have been proposed to deal with these problems in various aspects as in Table 1.

Table 1

The improvements of each aspect of CNN.

| Aspect | Literature | Contribution | Improvement |
|---|---|---|---|
| Convolutional layer | Ngiam(50) | Tiled CNN | It tiles and multiples feature maps to learn rotational and scale invariant features. |
| | Zeiler(34) | Transposed CNN | It can be seen as the backward pass of traditional CNNs. |
| | Yu(51) | Dilated CNN | It introduces more hyper-parameters to CNN which can increase receptive field size and cover more relevant information. |
| Pooling layer | Hyvärinen(52) | $L_P$ Pooling | It has a better generalization than max pooling. |
| | Zeiler(53) | Stochastic Pooling | It's a dropout-inspired pooling method which increases generalization of CNN. |
| | Yu(54) | Mixed Pooling | It can better address the overfitting problems. |
| | Gong(55) | Multi-scale Orderless Pooling | It improves the invariance of CNNs without degrading discriminative power. |
| | Rippel(56) | Spectral Pooling | Compared with max pooling, it can preserve more information for the same output dimensionality. |
| | He(57) | The second level headings | It can generate a fixed-length representation regardless of the input sizes. |
| Activation function | Nair(58) | ReLU | It is one of the most notable nonsaturated activation functions. |
| | Maas(59) | Leaky ReLU | It improves ReLU's disadvantage of having zero gradient. |
| | He(60) | Parametric ReLU | It reduces the risk of overfitting and improve the accuracy. |
| | Xu(61) | Randomized ReLU | It also reduces the risk of overfitting and improve the accuracy. |
| | Clevert(62) | ELU | It enables faster learning of DNNs and leads to higher classification accuracies. |
| | Goodfellow(63) | Maxout | It enjoys all the benefits of ReLU and it is well suited for training with dropout. |
| | Springenberg(64) | Probout | It can achieve the balance between preserving the desirable properties of maxout units and improving their invariance properties. |
| Loss function | Jin(65) | Hinge loss | It fasters the training and improves the accuracy. |
| | ---- | Softmax Loss | It is the combination of Multinomial Logistic loss and Softmax loss. |
| | Liu(66) | Large-Margin Softmax | It performs better than the Softmax. |
| | Lin(67) | Double Margin Loss | It improves the training accuracy. |
| | Schroff(68) | Triplet Loss | Its object is to minimize the distance between the anchor and positive, and maximize the distance between the negative and the anchor. |
| | ---- | Kullback-Leibler Divergence | It is widely used as a measure of information loss in the objective function of various Autoencoders. |
| Regularization | Tikhonov(69) | $l_2$-norm Regularization | It makes full use of the sparsity of weights to get a better optimization. |
| | Hinton(70) | $l_p$-norm Regularization | It makes the optimization easier. |
| | | Dropout | It is very effective in reducing overfitting. |
| | Wang(71) | Fast Dropout | It samples using Gaussian Approximation. |
| | Ba(72) | Adaptive Dropout | It reduces overfitting. |
| | Tompson(73) | Spatial Dropout | It reduces overfitting and is very suitable for the training of a small dataset size. |
| | Wan(74) | Drop Connect | It is used on the convolutional layers and reduces overfitting easier than dropout used on the full connection layers. |

### *2.3. Existing GAN models*

Generative image modeling is a fundamental problem in computer vision (75). There has been remarkable progress in this direction with the emergence of deep learning (DL) techniques. Variational AutoEncoder (VAE) (76, 77) formulates the problem with probabilistic graphical models whose goal is to maximize the lower bound of data likelihood. Autoregressive models, e.g. PixelRNN (78), which utilizes neural networks to model the conditional distribution of the pixel space have also generated appealing synthetic images. Recently, GAN (21) has shown a promising performance for generating sharper images.

The GAN is composed of generator $G(z)$ and discriminator $D(x)$. $G(z)$ maps a source of noise to the input space. $D(x)$ receives either a generated sample or a true data sample and must distinguish between them. Mathematically, the training process between $G(z)$ and $D(x)$ is the minimax objective (21)

$$\min_G \max_D V(D,G) = \mathbb{E}_{x\sim Pdata}\left[\log D(x)\right] + \mathbb{E}_{z\sim Pz}\left[\log\left(1-D\left(G(z)\right)\right)\right] \quad (11)$$

where $x$ is the image from training samples $P_{data}$, $z$ is a noise vector sampled from distribution $P_z$.

Conditional GAN (CGAN) (79) is an extension of GAN where both $G(z)$ and $D(x)$ receive additional conditioning variables $c$, yielding $G(z, c)$ and $D(x, c)$. This formulation allows $G(z)$ to generate images conditioned on variables $c$.

Training instability makes it hard for GAN to generate high-resolution photo-realistic images, the loss functions of $G(z)$ and $D(x)$ is unable to reflect the training process, and the generated images are lack of diversity. In order to solve those shortcomings, several GAN architectures, e.g. energy-based GAN (EBGAN) (80), metric learning-based GAN (MLGAN) (81), and deep convolutional GAN (DCGAN) (82) have been proposed and generated compelling results. DCGAN makes some significant improvements (82).

* **i.** Replace all deterministic spatial pooling functions, e.g. max pooling, using convolutional layer;
* **ii.** Add batch normalization in both $G(z)$ and $D(x)$ to stabilize learning;
* **iii.** Remove full connection layers to achieve deeper architectures and increase model stability;
* **iv.** Replace tanh activation function using ReLU and Leaky ReLU for $G(z)$ and $D(x)$ respectively to obtain a higher resolution modeling.

In order to improve the quality of generated images and stability of learning process. Mao (83) proposed Least Squares Generative Adversarial Network (LSGAN) which adopt the least squares loss function for the $D(x)$. The experimental results demonstrate that the LSGAN can generate more realistic images than regular GANs and have a more stability training process. Then the objective functions for the LSGAN can be defined as follows.

$$\max_D V(D) = \frac{1}{2}\mathbb{E}_{x\sim Pdata}\left[\left(D(x)-b\right)^2\right] + \frac{1}{2}\mathbb{E}_{z\sim Pz}\left[\left(D(G(z))-a\right)^2\right] \quad (12)$$

$$\max_G V(G) = \frac{1}{2}\mathbb{E}_{z\sim Pz}\left[\left(D(G(z))-c\right)^2\right] \quad (13)$$

where $a$=$c$=1, and $b$=0.

Although the improvements of DCGAN are effective, it is found the DCGAN seems to be powerless on high-resolution and high-complexion images. Therefore, Wasserstein GAN (WGAN), shown in Fig. 1, is proposed and several simple effective are improved (84).

* **i.** Remove the Sigmoid layer in $D(x)$;
* **ii.** Remove the logarithmic algorithm of loss functions;
* **iii.** The update scope of parameters in $D(x)$ is fixed as a constant;
* **iv.** Momentum-based optimization algorithm is not recommended, while root mean square prop (RMSProp) and stochastic gradient descent (SGD) are.

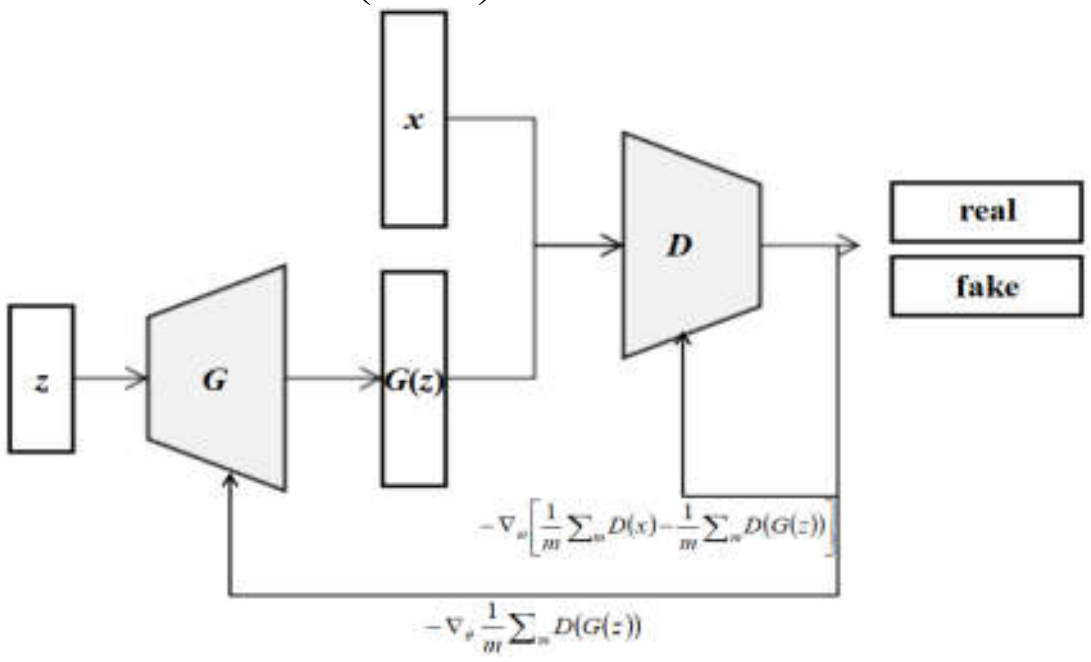

Fig. 1. The structure of WGAN.

$\omega$ is parameters of $D(x)$, $\theta$ is parameters of $G(z)$, $m$ is the batch size.

Then, Ishaan Gulrajani et al. (85) pointed out that although the WGAN makes progress toward stable training, it sometimes still generates only poor samples or fail to converge. This problem mainly due to the use of weight clipping in the WGAN to enforce a Lipschitz constraint on the discriminator. Therefore, they proposed gradient penalty to enforce the Lipschitz constraint. The new optimization objective is

$$L = \underbrace{\mathbb{E}_{x\sim Pdata}\left[D(x)\right] - \mathbb{E}_{z\sim Pz}\left[D(G(z))\right]}_{\text{Original discriminator loss}} + \underbrace{\lambda\mathbb{E}\left[\left(\left\|\nabla_x D(x)\right\|_2 - 1\right)^2\right]}_{\text{Gradient penalty proposed by WGAN-GP}} \quad (14)$$

## 3. Reconstruction neural network

### *3.1. Input data*

In this study, the topology optimization is employed as an experimental example and the training samples of the network are the contour images which are shown in Fig. 2, and each image's pixel is 469×469. The samples contain 22000 contour images and each label is the compliance. The topology opti-mization problem based on the power-law approach, where the objective is to minimize the compliance can be written as (86)

$$\min_{\mathbf{x}} : c(\mathbf{x}) = \mathbf{U}^T\mathbf{K}\mathbf{U} = \sum_{e=1}^{N} (x_e)^p \mathbf{u}_e^T \mathbf{k}_0 \mathbf{u}_e \tag{15}$$

subject to

$$\frac{V(\mathbf{x})}{V_0} = f \tag{16}$$

$$\mathbf{K}\mathbf{U} = \mathbf{F} \tag{17}$$

$$\mathbf{0} < \mathbf{x}_{\min} \le \mathbf{x} \le \mathbf{1} \tag{18}$$

where **U** and **F** are the global displacement and force vectors, respectively, **K** is the global stiffness matrix, $\mathbf{u}_e$ and $\mathbf{k}_0$ are the element displacement vector and stiffness matrix, respectively, **x** is the vector of design variables, $\mathbf{x}_{\min}$ is a vector of minimum relative densities (non-zero to avoid singularity), N (=nelx×nely) is the number of elements used to discretize the design domain, $p$ is the penalization power (typically $p$=3), $V(\mathbf{x})$ and $V_0$ is the material volume and design domain volume, respectively, and f(volfrac) is the prescribed volume fraction.

Considering each image is symmetric, the upper part of each images is trimmed and the useless blank around the structural configuration is cut off in order to reduce the size of input. Actually, each sample's pixel shown in Fig. 3 is 117×390×1 (monochrome image).

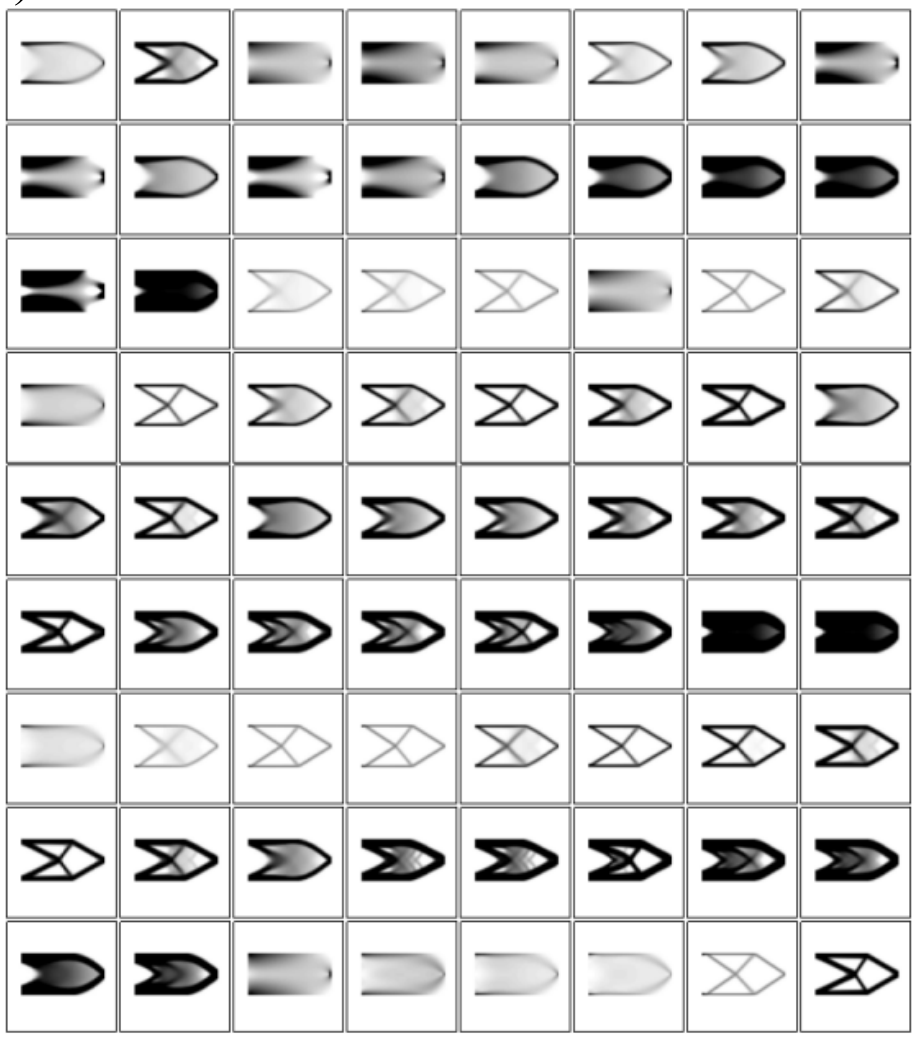

Fig. 2. Original training samples.

According to the class histogram of labels shown in Fig. 4a whose abscissa is the range of compliance and ordinate is the count of samples, most labels are concentrate in [0, 150], and few of them are in [150, 500]. Obviously, it is not a suitable distribution for the network's training. Therefore, the labels are normalized by three strategies and their mathematical description are presented as follows.

(1) **Min-max normalization**

$$y' = \frac{y - y_{\min}}{y_{\max} - y_{\min}} \tag{19}$$

where $y'$ is the normalized label, $y$ is the label, $y_{\max}$ is the maximum label, and $y_{\min}$ is the minimum label.

(2) **Z-score normalization**

$$y' = \frac{y - \mu}{\sigma} \tag{20}$$

where $\mu$ is the mean value of labels, and $\sigma$ is the standard deviation of labels.

(3) **Logarithm normalization**

$$y' = \log_C y \tag{21}$$

where $C$ is a constant. In this study, $C$=20.

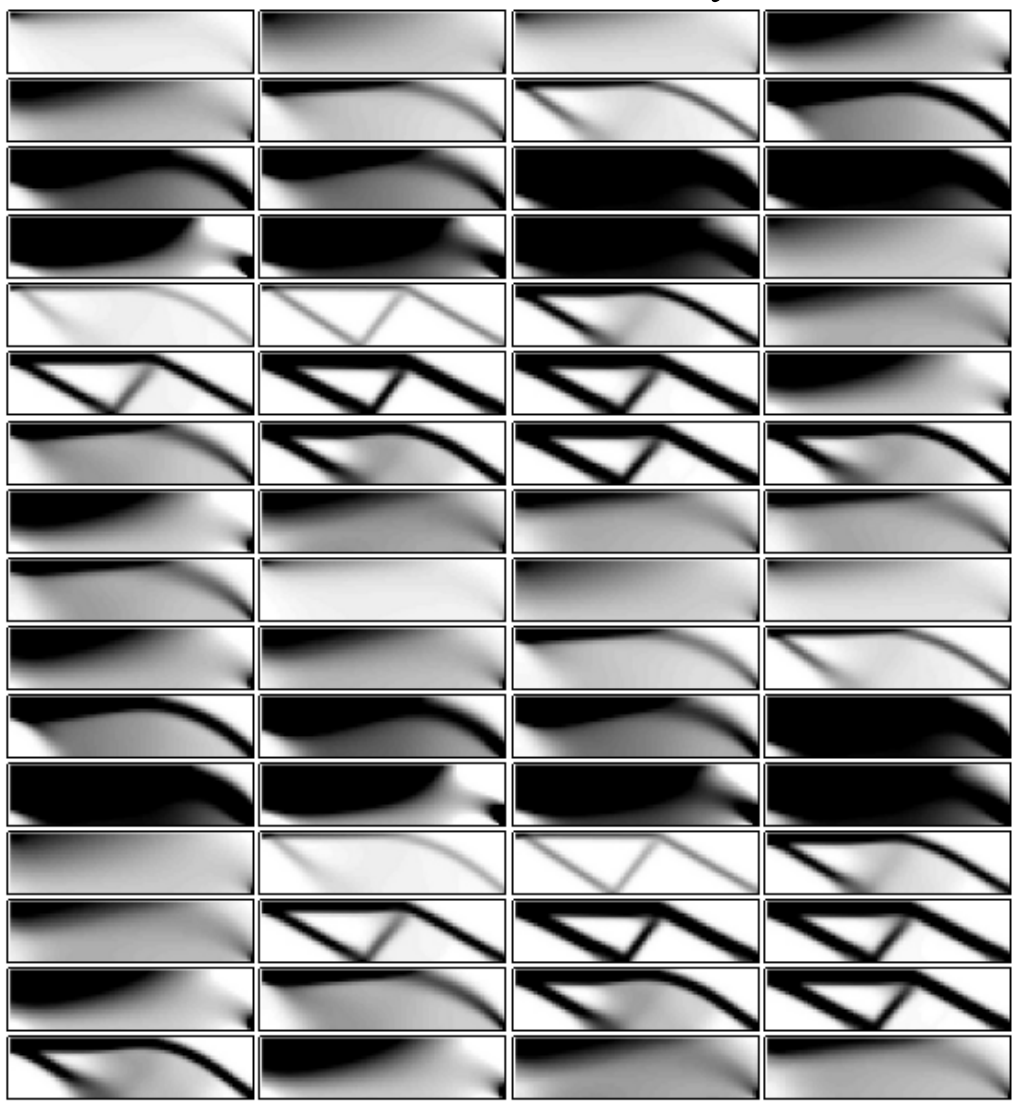

Fig. 3. Actual training samples.

The normalization results are shown in Figs. 4(b-d). It can be found that both the interval lengths of normalization utilizing min-max and logarithm are approximate to 1 and Z-score normalization's interval length is 6.5. The distribution of normalized data associated with min-max and Z-score are similar to the original data distribution, and logarithm normalization makes the original data more uniform in the entire space. Therefore, the logarithm optimization is employed for easier network training.

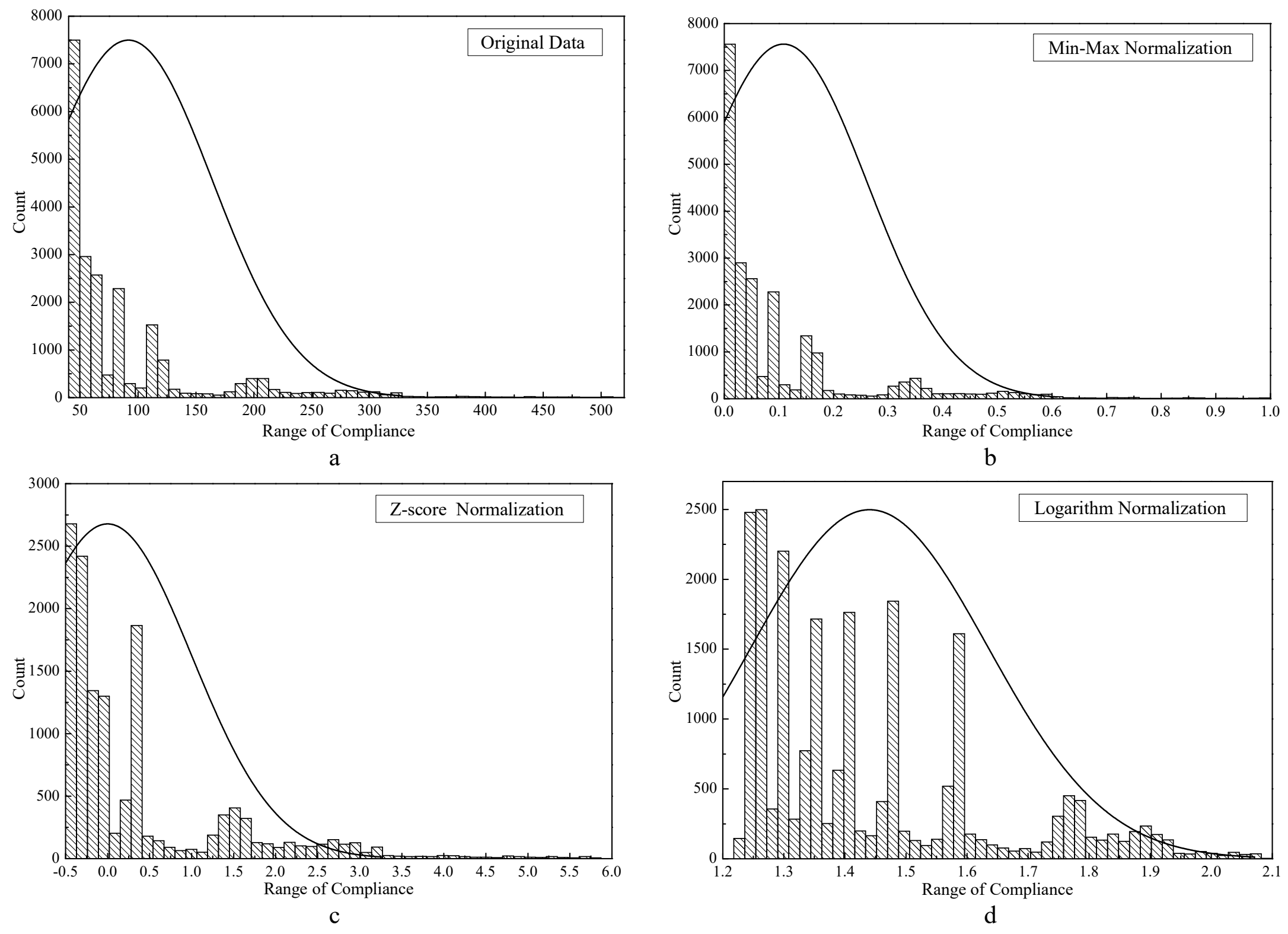

Fig. 4. Comparison of class histograms of normalized labels by different strategies.

### 3.2. Convolution in Convolution submode

To generate high-accuracy regression results, an effective CIC submodel is proposed. As shown in Fig. 5, the images-to-regression process is decomposed into three stages.

- **Stage-I CIC:** The original image is divided into 30 subimages of 39×39×1;
- **Stage-II CIC:** Each subimage is trained using a separate subCNN which has three convolutional and three pooling layers. The output size is 1×1×1;
- **Stage-III CIC:** All outputs are collected from subCNNs and reshaped into a matrix of 3×10×1. The matrix is transmitted into the full connection layer, which has one convolutional and one pooling layers.

Compared with the classical convolutional layer, if the input size is 117×390×1, the kernel size is 5×5 and kernel deep is 32, then the number of this layer's parameters is 5×5×1×32+32. As for the CIC, its convolutional layer has 30×(5×5×1×32+32) parameters for 30 subCNNs under the same condition. While, the accuracy of the CIC can be improved through 30 times the parameters of classical convolutional layer without increasing the cost of computation because each subCNN's input is far smaller than the original image. The optimization algorithm used in the CIC is adaptive moment estimation (Adam) optimizer which is essential RMSProp with momentum factor and its advantages are as follows.

- **i.** Adam combines the advantages of Adagrad's ability to handle sparse gradients and RMSProp's ability to deal with non-stationary targets;
- **ii.** Adam spends lower computational cost;
- **iii.** Adam can calculate different adaptive learning rates for different parameters;
- **iv.** Adam performs well for most nonconvex optimization, large data sets, and high-dimensional space.

Mathematically, the Adam can be expressed as

$$g \leftarrow +\frac{1}{m}\nabla_{\theta}\sum_{i}L\left(f\left(x_i;\theta\right),y_i\right) \tag{22}$$

$$s \leftarrow \rho_1 s + (1-\rho_1) g \tag{23}$$

$$r \leftarrow \rho_2 r + (1-\rho_2) g \odot g \tag{24}$$

$$\hat{s} \leftarrow \frac{s}{1-\rho_1} \tag{25}$$

$$\hat{r} \leftarrow \frac{r}{1-\rho_2} \tag{26}$$

$$\Delta\theta = -\varepsilon \frac{\hat{s}}{\sqrt{\hat{r}} + \delta} \tag{27}$$

$$\theta \leftarrow \theta + \Delta\theta \tag{28}$$

where $\theta$ is the initial parameters, $x_i$ is the training samples and $y_i$ is corresponding labels, $m$ is the number of samples, $s$ and $r$ are the first and second moment estimation respectively, $\rho$ is the attenuation coefficient, and $\varepsilon$ is the learning rate. In this study, $\delta$=$10^{-8}$, $\rho_1$=0.9, and $\rho_2$=0.999.

The algorithm is described in Table 2.

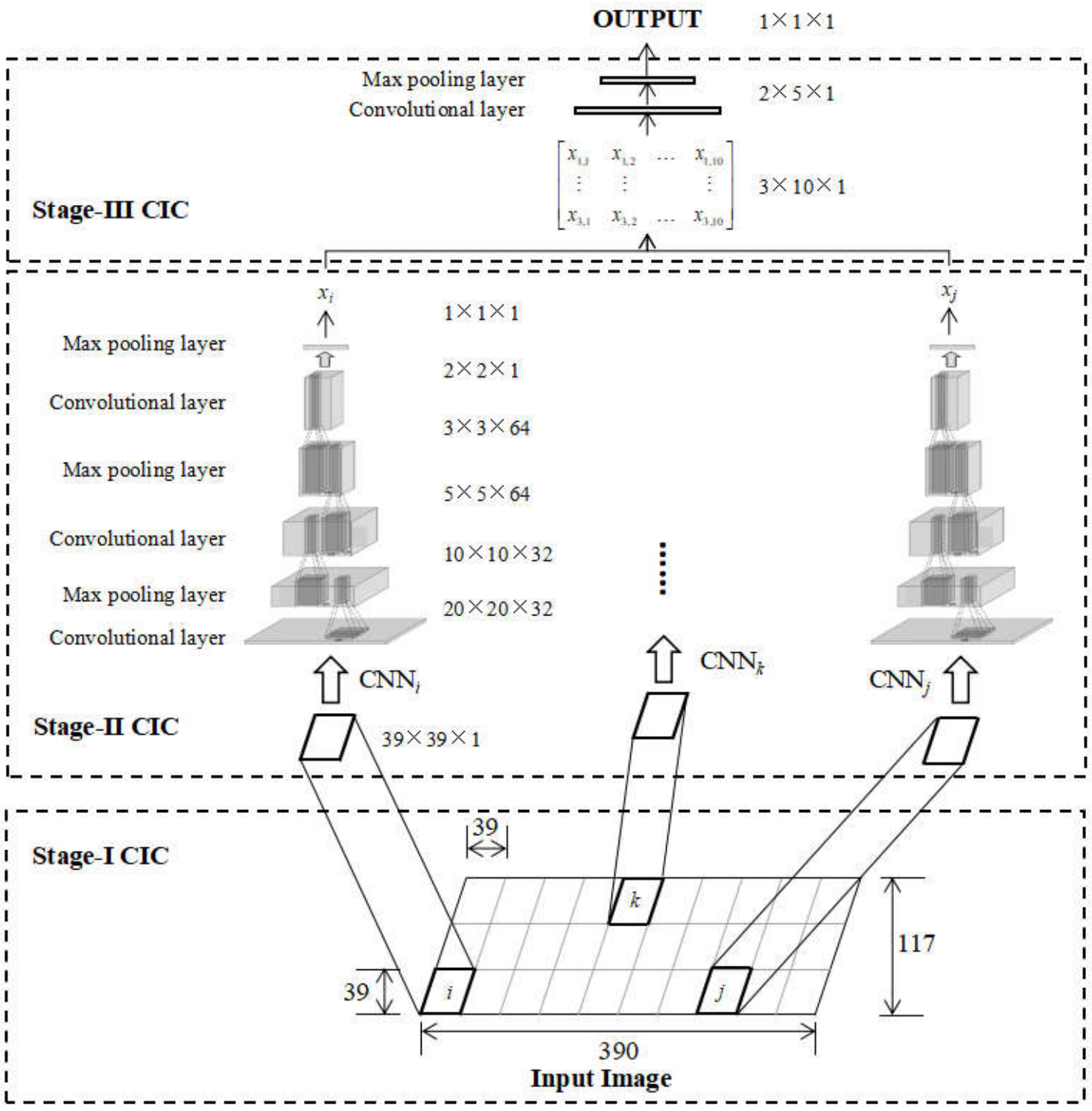

Fig. 5. The proposed CIC model.

Table 2

The algorithm of CIC submodel.

| |
|---|
| **Algorithm** CIC, our proposed algorithm. All experiments in this study used the default values $\alpha_0 = 0.01$, $\triangle = 0.99$, $d = 50$, $m = 50$, $\eta = 0.0001$, $t = 5000$. |
| **Require:** $\alpha$, the learning rate. $\alpha_0$, the initial learning rate. $\triangle$, the attenuation coefficient of learning rate. $m$, the batch size. $\eta$, the proportion of l2 regularization to the loss. $t$, the training steps.<br>**Require:** $w_0$, initial weight parameters. $b_0$, initial bias parameters.<br>1: **for** $i = 0, \ldots, t$ **do**<br>2: **while** *loss* has not converged **do**<br>3: Sample $\{x^{(t)}\}_{t=1}^{m}$ a batch from the training samples. |

$loss \leftarrow \frac{1}{m}\sum_{t=1}^{m}\left(CIC\left(x^{(t)}\right)-label^{(t)}\right)^2+\eta\cdot\sum_i\left|w_i^2\right|$

$\alpha \leftarrow \alpha_0 \cdot \triangle$ ^ ( $i$ / $d$)

$w \leftarrow w - \alpha \cdot$ AdamOptimizer (*loss*)

$b \leftarrow b - \alpha \cdot$ AdamOptimizer (*loss*)

**end while**

**end for**

### *3.3. Convolutional AutoEncoder based on Wasserstein Generative Adversarial Network*

As mentioned above, existing GAN models still face many unsolved difficulties, and through considerable amount of experimental work, existing models appear powerless to generate satisfied contour images during this study. To enhance the training stability, reduce the training difficulty, and improve the training accuracy, the CAE is employed and the WGAN-CAE submodel shown in Fig. 6 is proposed. Through the CAE, the input size of the WGAN can be reduced to 8×8. It will greatly improve the accuracy and reduces the training difficulty of the WGAN due to reduction of input size. The WGAN-CAE structures are presented in Table 3. To clarify, the ReConNN model is summarized in Fig.6. The optimization algorithm used in the CAE is Adam optimizer and used in the WGAN is the root mean square prop (RMSProp) optimizer and its mathematical expression is

$$g \leftarrow +\frac{1}{m}\nabla_\theta \sum_i L\left(f\left(x_i;\theta\right), y_i\right) \tag{29}$$

$$r \leftarrow \rho r + \left(1-\rho\right) g \odot g \tag{30}$$

$$\Delta\theta = -\frac{\varepsilon}{\delta+\sqrt{r}} \odot g \tag{31}$$

$$\theta \leftarrow \theta + \Delta\theta \tag{32}$$

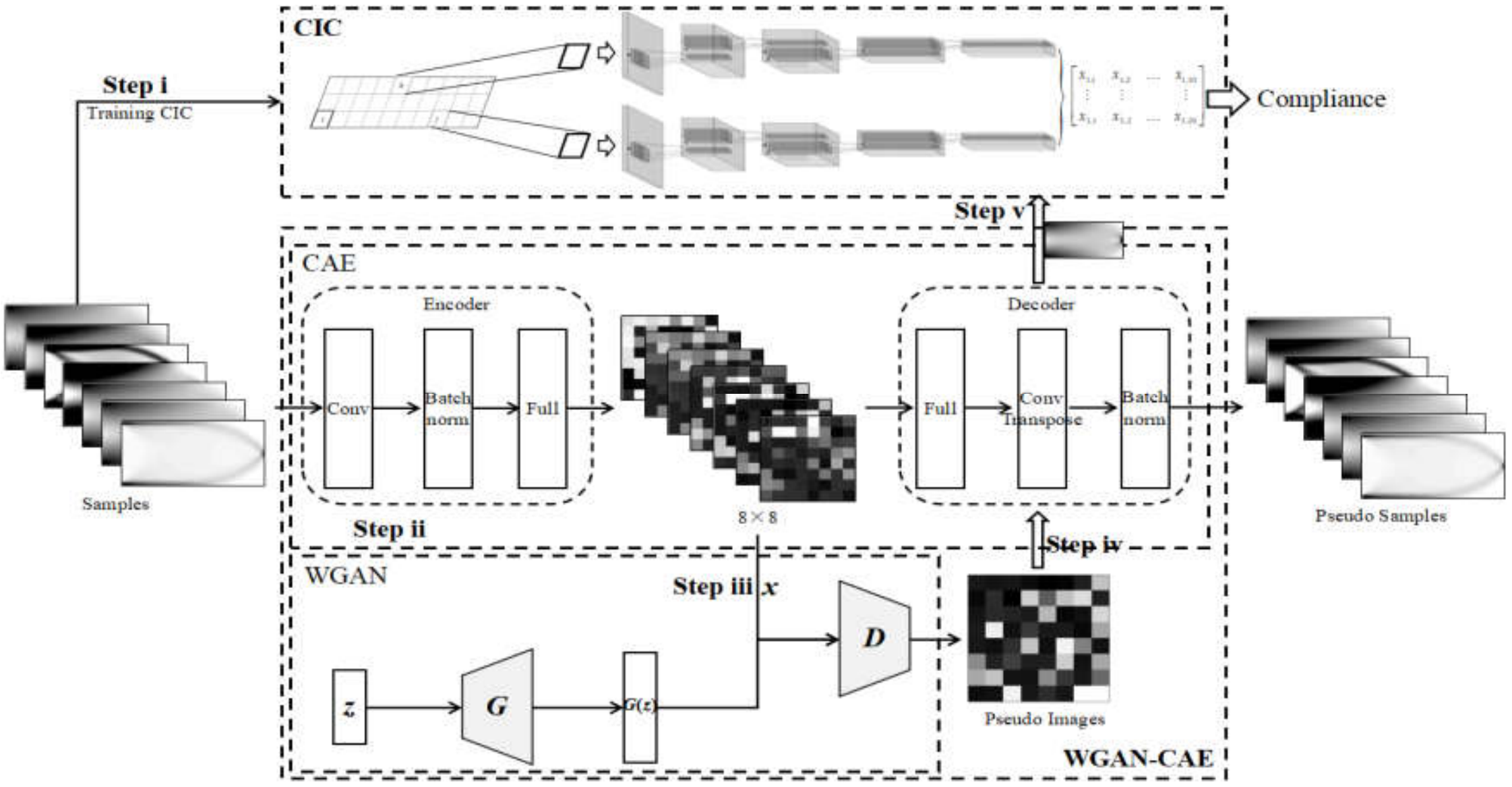

Fig. 6. The proposed ReConNN model.

The ReConNN model contains two submodels. The CIC submodel constructs the mapping between contour images and objective function using existing samples in the Step **i**. The WGAN-CAE submodel, which is integrated of WGAN and CAE, is employed to generate more "pseudo images" in the Step **ii**. Among the WGAN-CAE, the encoder of CAE compresses the samples to 8×8 and the decoder restores the compressed samples to the original size. On the other hand, the WGAN generates pixel images which are similar to compressed samples in the Step **iii** and the "pseudo images" are restored using trained decoder in the Step **iv**. Finally, in the Step **v**, each objective function of restored "pseudo sample" is calculated by trained CIC.

Table 3

Architecture parameters of WGAN-CAE.

| | Model | Network Architecture | Activation Function | Optimization Algorithm |
|---|---|---|---|---|
| CAE | $E(y)$ | 3 convolutional and 4 full connection layers. | Leaky ReLU | Adam Optimizer |
| | $D(x)$ | 4 full connection and 3 deconvolutional layers. | Leaky ReLU | |
| WGAN | $G(z)$ | 1 full connection, 2 deconvolutional, and 1 batch norm layers. | ReLU | RMSProp Optimizer |
| | $D(x)$ | 2 convolutional, 1 batch norm layer, and 1 full connection layers. | Leaky ReLU | |

## 4. Experiments and Analysis

In this study, the classical topology optimization is employed as an experimental example to test the proposed ReConNN model. The input samples are the contour images shown in Fig. 3 and each sample's label is the compliance calculated using Eqs. (15)-(18).

To validate performances of the CIC and WGAN-CAE, quantitative and qualitative evaluations are conducted. Three conditional CNN models whose structures are shown in Table 4 on images-to-classification, LeNet-5 (87), AlexNet (33), and VGG (35), and three high-performance GAN models shown in Table 5, DCGAN (82), WGAN (84) and WGAN-GP (85), are compared. ReLU and Leaky ReLU (LReLU) functions are calculated by Eqs. (33) and (34) respectively.

$$\text{ReLU}=\max\left(x,0\right) \tag{33}$$

$$\text{LReLU}=\max\left(x,\lambda x\right) \tag{34}$$

where $\lambda$ is a random number between 0 and 1. In this study, $\lambda$=0.2.

### 4.1. Performance criteria

In order to evaluate the accuracy of these CNNs, three criteria (1, 8, 88) in Table 6 are employed. The RMSE indicates the overall accurate approximation, while the MAE reveals the presence of regional areas of poor approximation.

In addition, the following relative error is used to evaluate the accuracy of model. In order to shown the relative error clearly, the (1-Error) is calculated in the later sections.

$$\text{Error}=\frac{\left\|\hat{y}-y\right\|_2}{\left\|y\right\|_2}\times 100\% \tag{35}$$

For generative models (e.g. GAN), it is difficult to evaluate the performance. Therefore, a recently proposed numerical assessment approach "inception score" (89) for quantitative evaluation is employed.

$$I=\exp\left(\mathbb{E}_x D_{KL}\left(p\left(y\mid x\right)\|\, p\left(y\right)\right)\right) \tag{36}$$

where $x$ denotes one sample, $p(y|x)$ is the softmax output of a trained classifier of the labels, and $p(y)$ is the overall label distribution of generated samples.

The intuition behind this criterion is that a good model should generate diverse but meaningful images. Therefore, the KL divergence between the marginal distribution $p(y)$ and the conditional distribution $p(y|x)$ might be large.

### 4.2. Results and discussions

#### 4.2.1. Performances of different CNN models.

As shown in Fig. 7, more than 80% (1-Error)s of CIC and AlexNet are larger than 80%. It can be found that CIC and AlexNet are more accurate than others on the contour images regression.

In order to provide a more comprehensive criterion of each CNN's ability, the number of samples trained per second, CPU utilization percentage, and other three criteria listed in Table 7 are employed. Considering the number of samples trained per second and CPU utilization percentage, the CIC has an obvious advantage. As for other three criteria, the AlexNet performs best.

To further analyze the results of CIC and AlexNet, the labels and predicted values of CIC and AlexNet are shown in Fig. 8. Surprisingly, although the AlexNet performs better than CIC in terms of accuracy and errors, the AlexNet's output trends to constant. As shown in Figs. 9 - 10, each abscissa indicates the class histogram which show the distribution of weights and biases, and ordinate indicates the training steps. Each left figure is the change process of biases during training, and the right is the weights' change process. As shown in Fig. 9, the AlexNet's weights and biases are constant expect the 3rd convolutional layer's bias starting from the second training step. It suggests that the metamodel constructed by the AlexNet only depends on a layer's biases to determine the output and such a metamodel is considered to be a bias one. Meanwhile, as shown in Fig. 10, weights and biases of all sub-CNNs and the full connection layer have a certain training tendency during the training process. It illustrates that the metamodel constructed by the CIC might represent the essential of the assigned problem.

Table 4

Architecture parameters of conditional CNNs.

| | Network Architecture | Activation Function | Optimization Algorithm |
|---|---|---|---|
| AlexNet | 5 convolutional and 3 full connection layers | | |
| LeNet | LeNet-5 | ReLU | Adam Optimizer |
| VGG | VGG16 | | |

Table 5

Architecture parameters of high-performance GANs.

| Model | | Network Architecture | Activation Function | Optimization Algorithm |
|---|---|---|---|---|
| DCGAN | $G(z)$ | 4 deconvolutional, 4 batchnorm and 1 full connection layers | ReLU | Adam Optimizer |
| | $D(x)$ | 4 convolutional, 4 batchnorm and 1 full connection layers | Leaky ReLU | |
| WGAN | $G(z)$ | 4 deconvolutional, 4 batchnorm and 1 full connection layers | ReLU | RMSProp Optimizer |
| | $D(x)$ | 4 convolutional, 4 batchnorm and 1 full connection layers | Leaky ReLU | |
| WGAN-GP | $G(z)$ | 4 deconvolutional, 4 batchnorm and 1 full connection layers | ReLU | RMSProp Optimizer |
| | $D(x)$ | 4 convolutional, 4 batchnorm and 1 full connection layers | Leaky ReLU | |

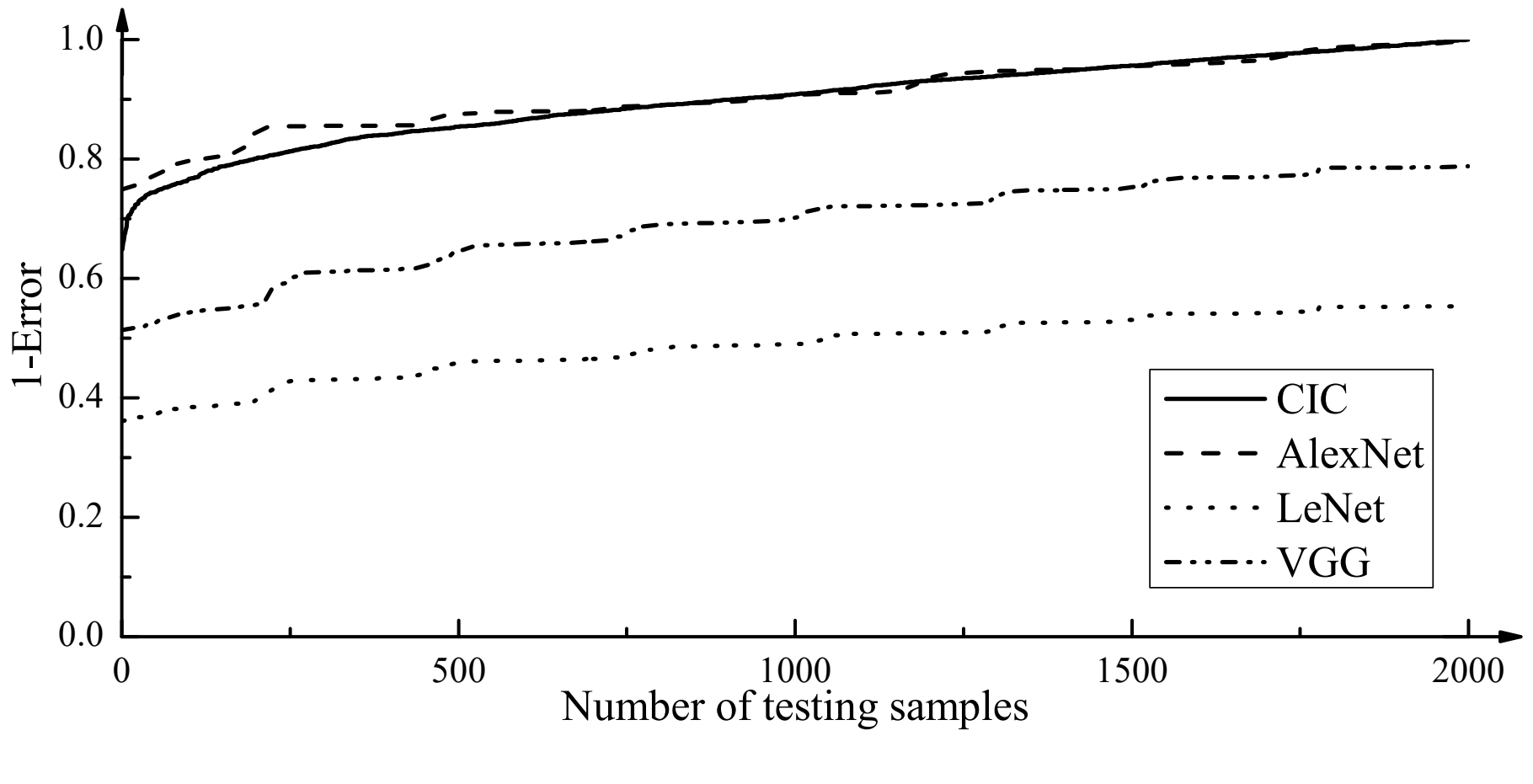

Fig. 7. Training result of different CNN models.

Table 6

Criteria for performance evaluation.

| Criteria | Expression |
|---|---|
| Maximal absolute error (MAE) | $\max\left(\lvert y_i - \hat{y}_i \rvert\right)$ |
| Average absolute error (AAE) | $\left(\sum_{i=1}^{n}\left\lvert y_i-\hat{y}_i\right\rvert\right)\Big/ n$ |
| Root mean square error (RMSE) | $\sqrt{\sum_{i=1}^{n}\left(y_i-\hat{y}_i\right)^2}\Big/\sqrt{n}$ |

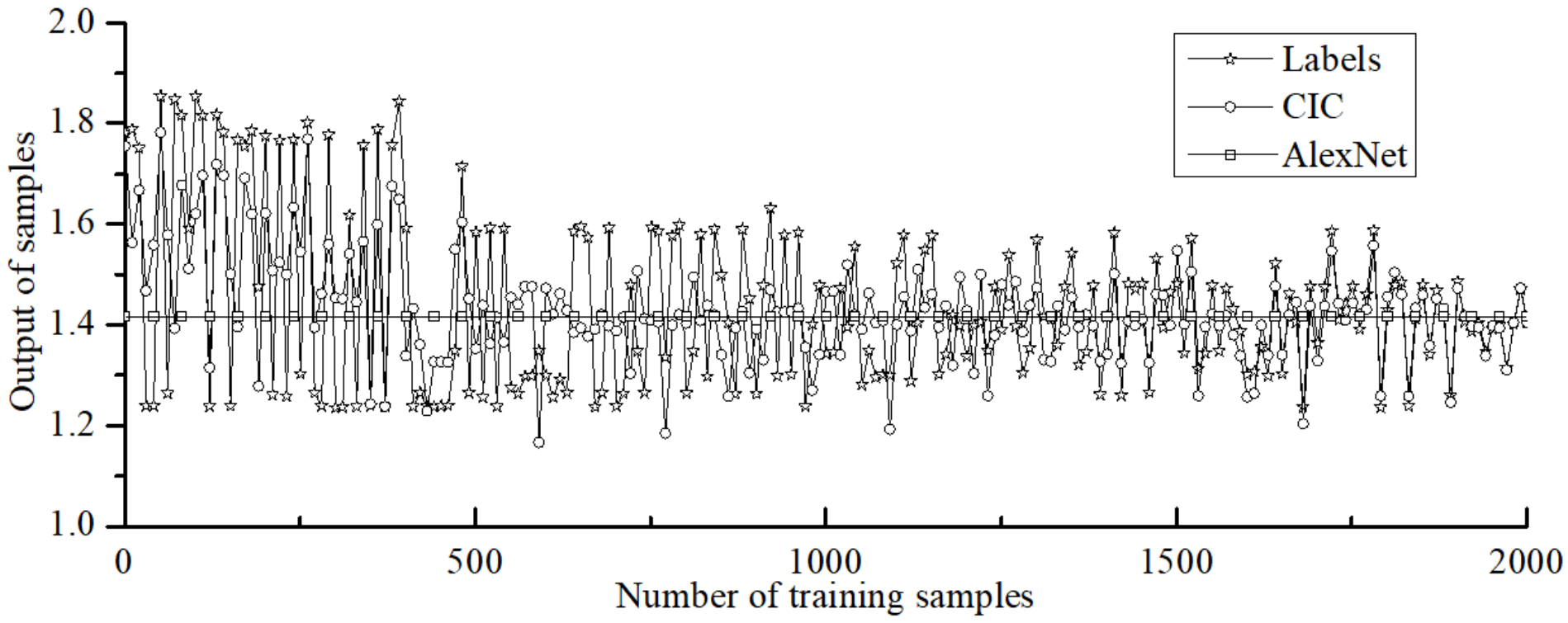

Fig. 8. Comparison of CIC and AlexNet.

Table 7

Criteria between different CNN models.

| Model | Time *samples/sec* | CPU Utilization Percentage | MAE | AAE | RMSE |
|---|---|---|---|---|---|
| CIC (ours) | 76.6 | 25.1% | 0.6258 | 0.1481 | 0.0354 |
| AlexNet | 77.6 | 32% | 0.4746 | 0.1343 | 0.0278 |
| LeNet | 70 | 21.4% | 1.2099 | 0.7397 | 0.5755 |
| VGG | 14.4 | 72% | 0.9217 | 0.4527 | 0.2346 |

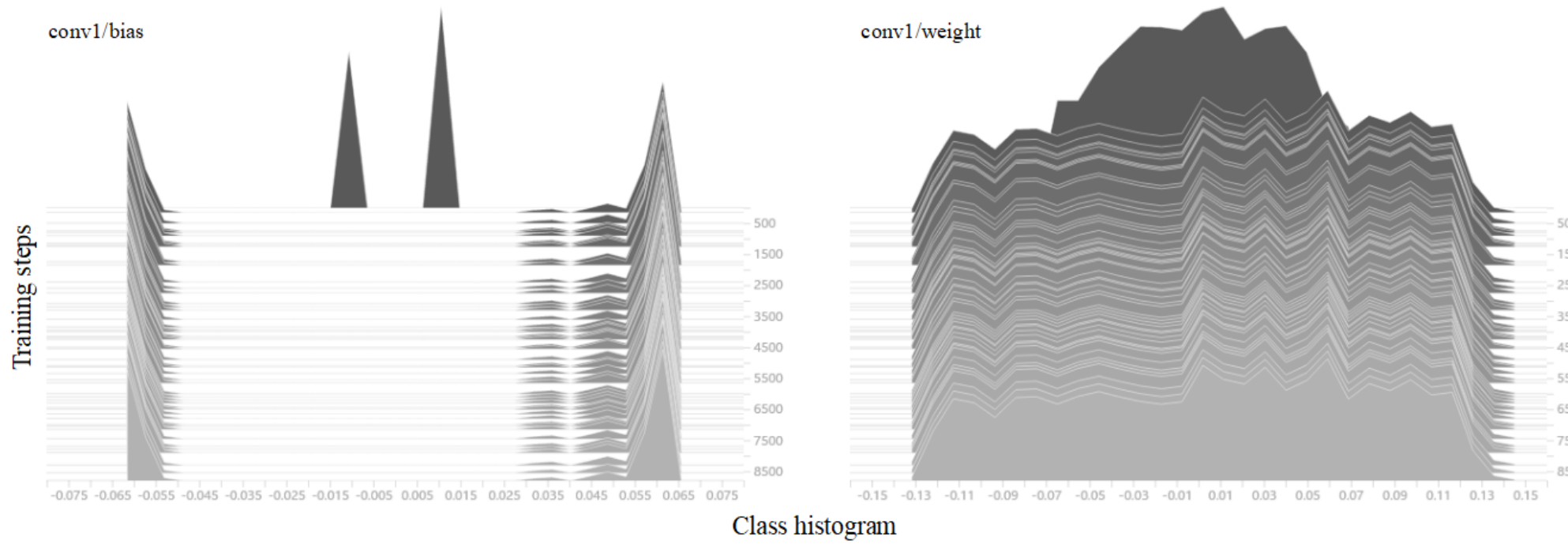

a. The 1st convolutional layer.

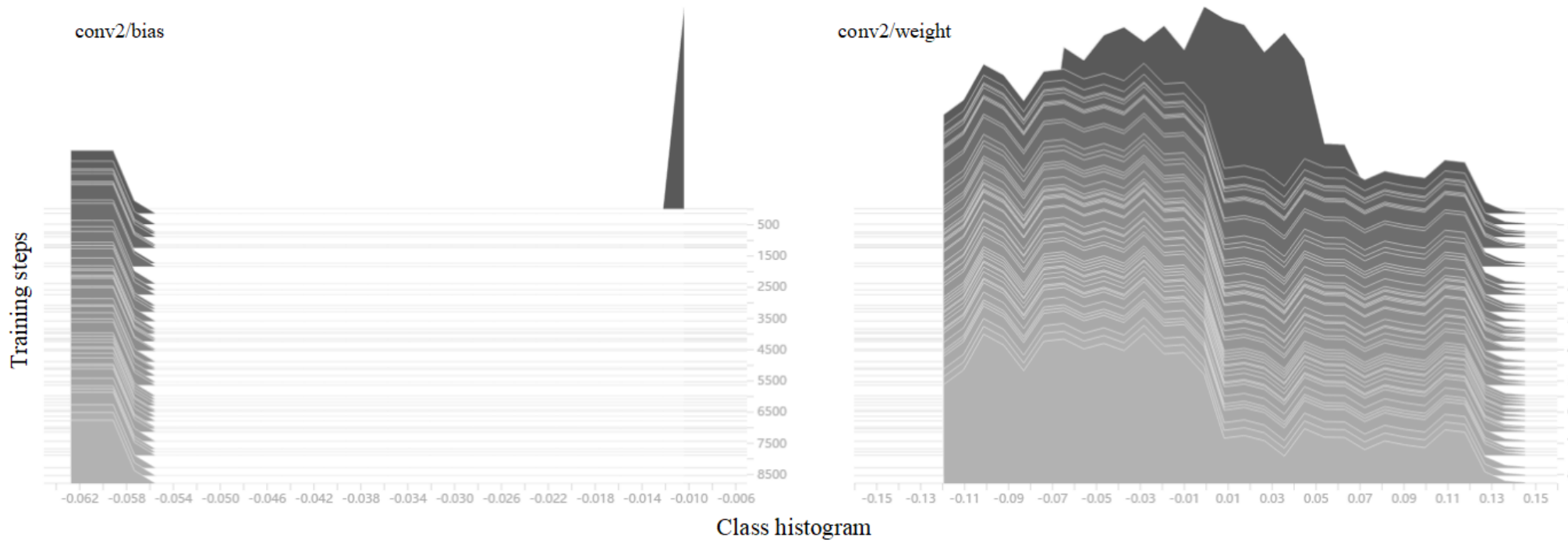

b. The 2nd convolutional layer.

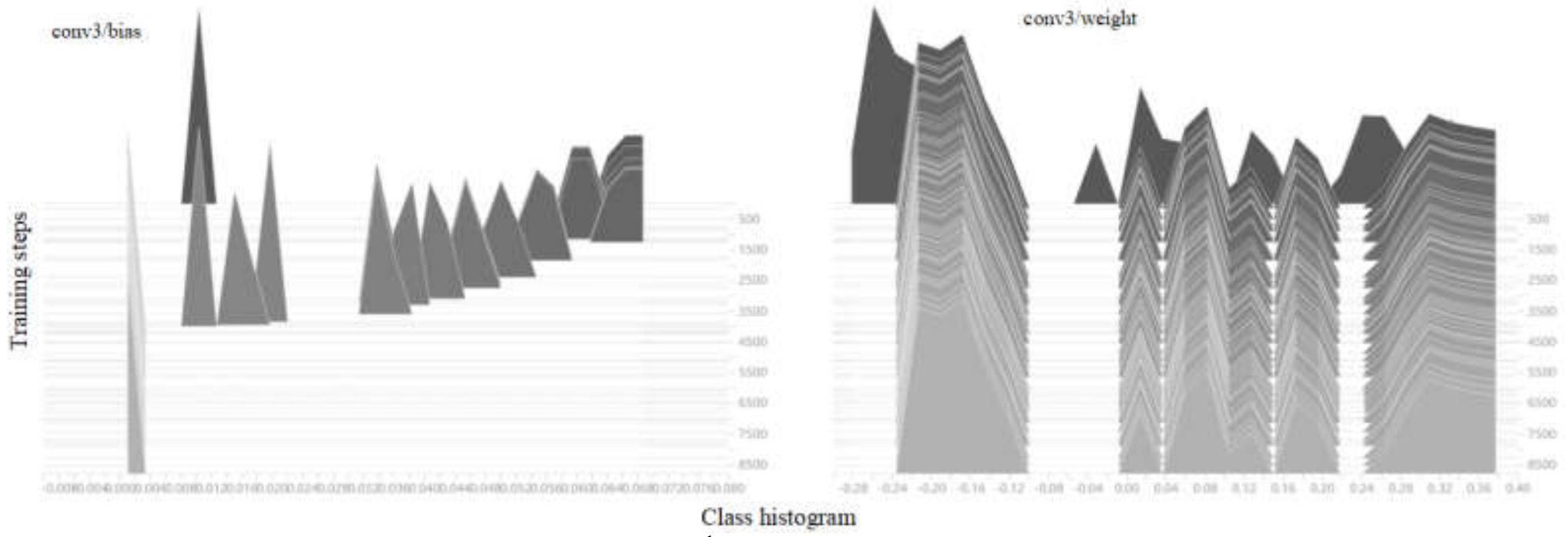

c. The 3rd convolutional layer.

Fig. 9. Training process of AlexNet.

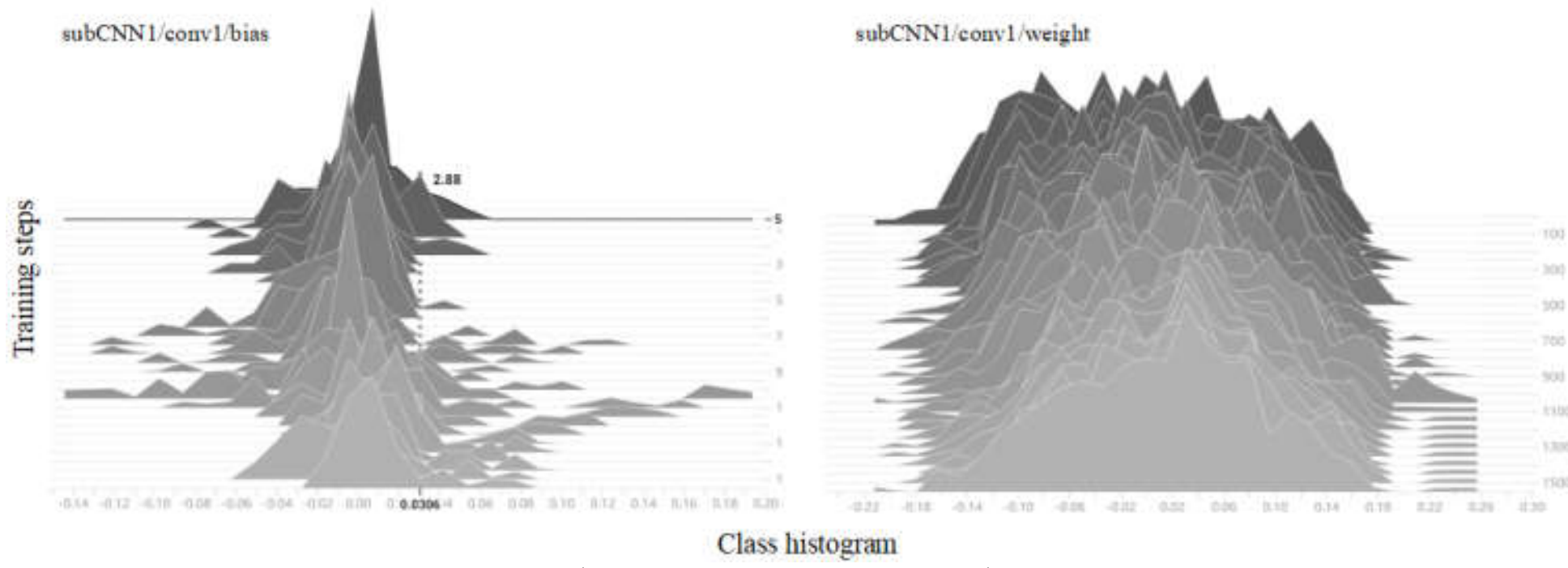

a. The 1st convolutional layer of the 1st subCNN.

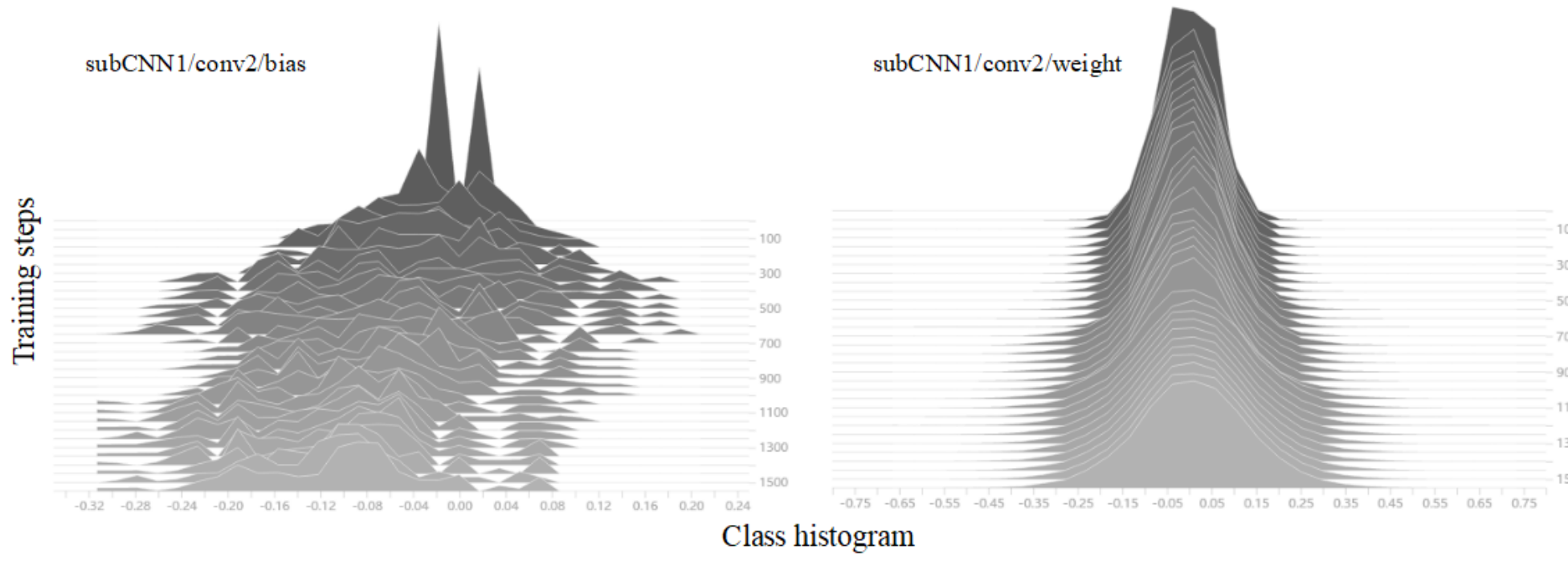

b. The 2nd convolutional layer of the 1st subCNN.

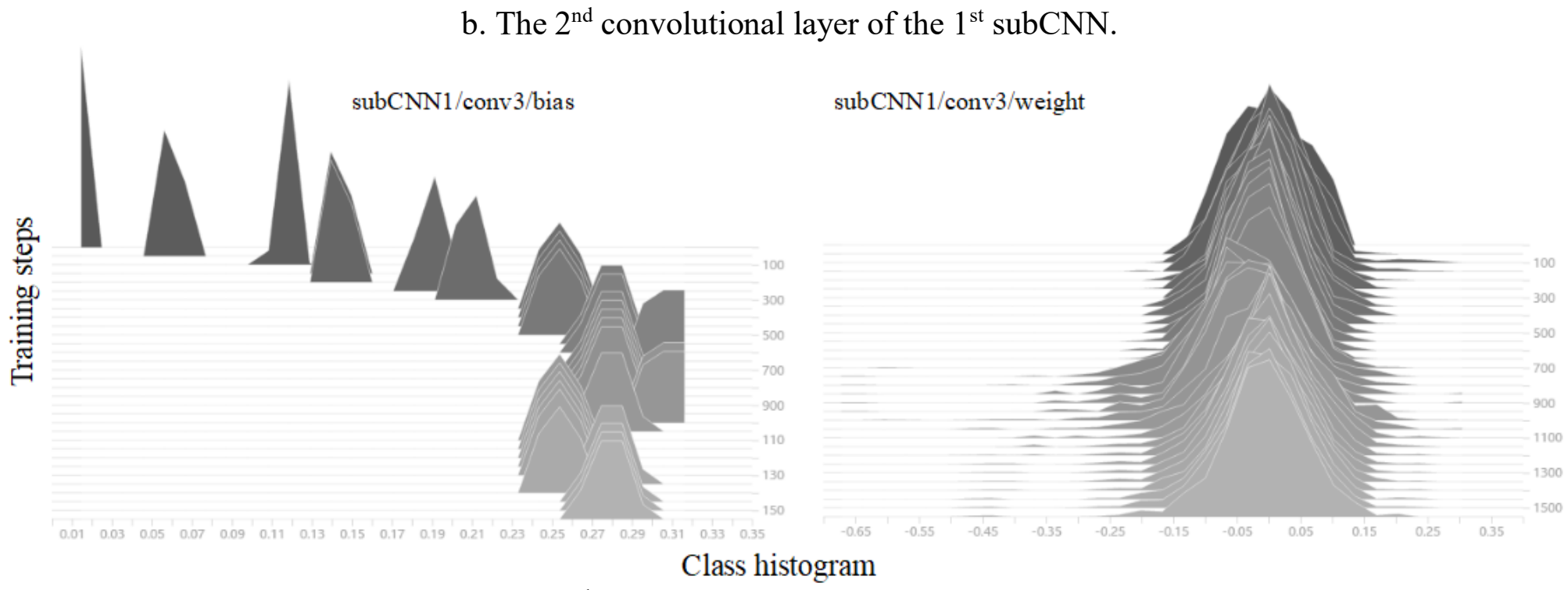

c. The 3rd convolutional layer of the 1st subCNN.

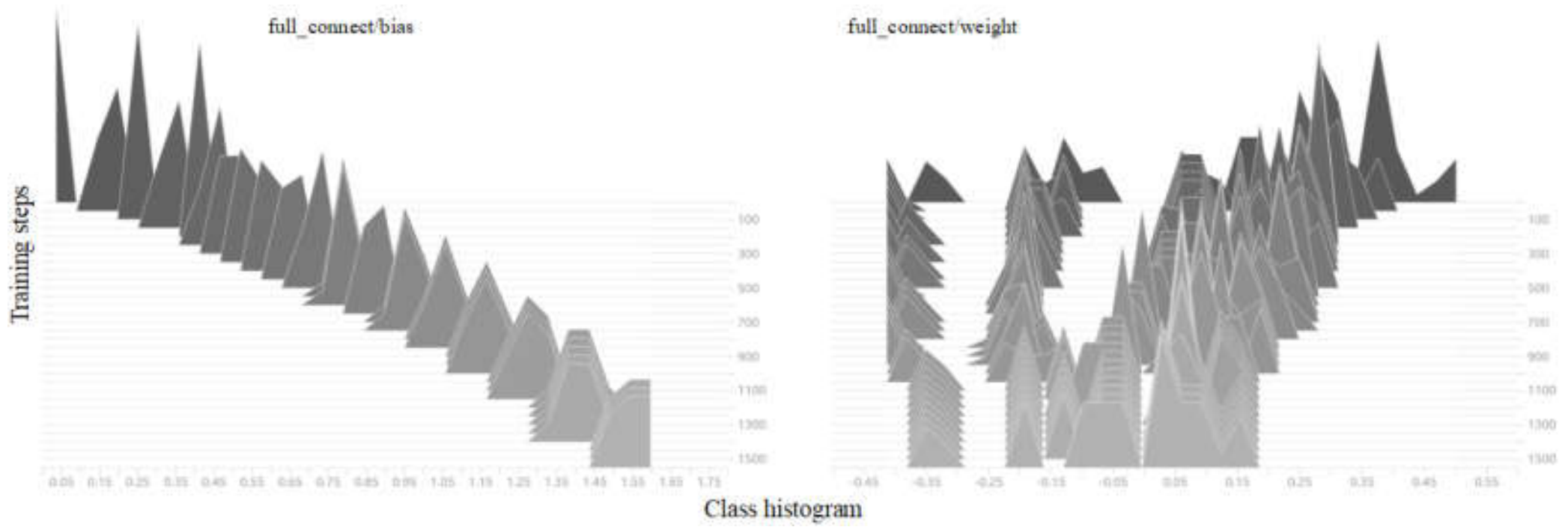

d. The full connection layer.

Fig. 10. Training process of CIC.

*4.2.2. Performances of different GAN models.*

Firstly, the CAE is trained and the loss (mean square error, MSE) (90) calculated by Eq. (37) and the training results are shown in Fig. 11 and Table 8. Then the trained $E(y)$ is applied to compress the samples for WGAN and the compressed image size is 8×8. The *g_loss* of WGAN's $G(z)$ during training process is shown in Fig. 12 which can be evaluated by Eq. (38).

$$MSE = \frac{1}{m \cdot n} \sum_{i=1}^{m} \sum_{j=1}^{n} \left( p_{i,j} - \hat{p}_{i,j} \right)^2 \tag{37}$$

where $m$ is the number of samples, $n$ is the number of pixels of each sample, and $p_{i,j}$ and $\hat{p}_{i,j}$ are the $j$-th pixel and predicted pixel of $i$-th sample respectively.

$$g_loss = \frac{1}{m} \left| \sum_{m} D(x) - \sum_{m} D(G(z)) \right| \tag{38}$$

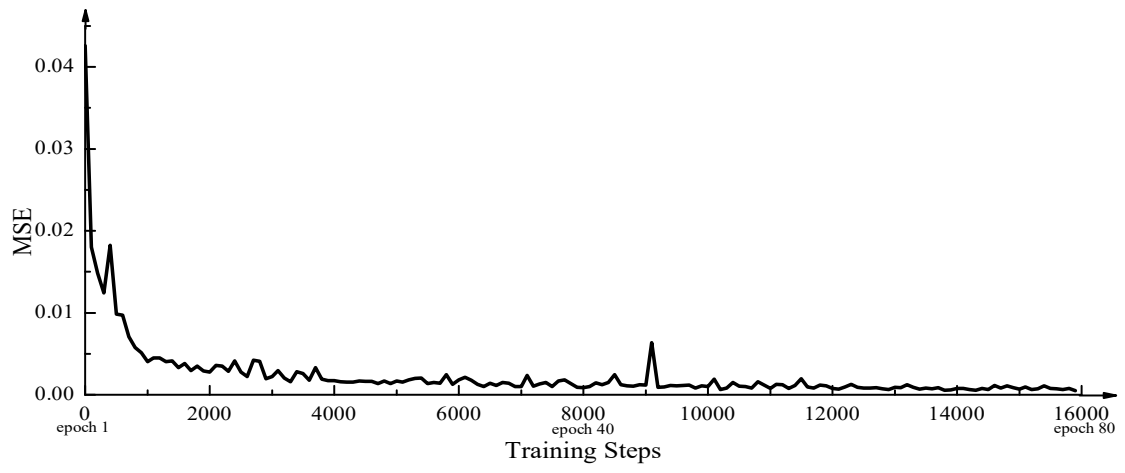

Fig. 11. The MSE during CAE's training.

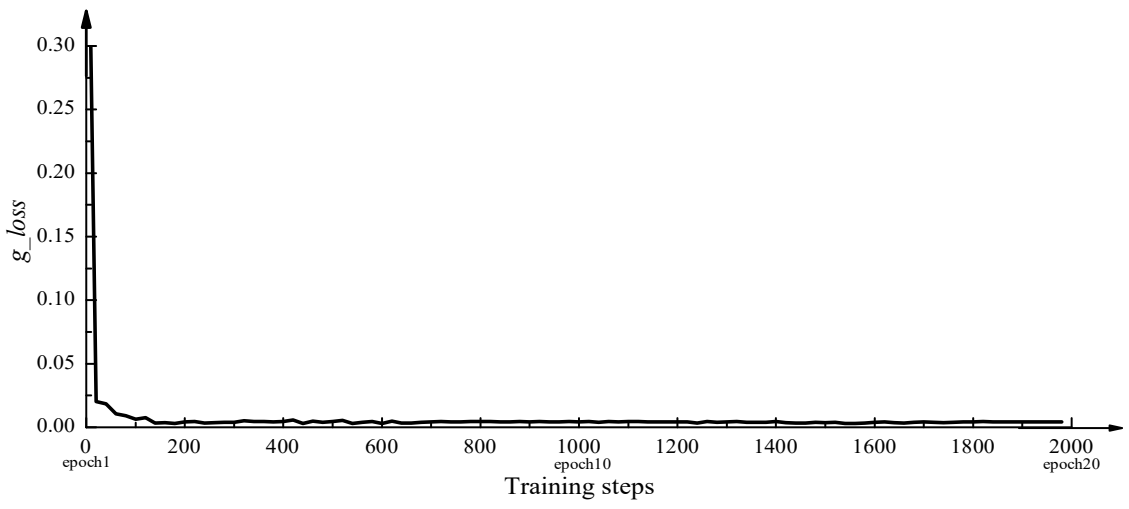

Fig. 12. The *g_loss* of $G(z)$ during WGAN's training.

Table 8

Comparison of input and predicted images using trained CAE.

| | |
|---|---|
| Original images | 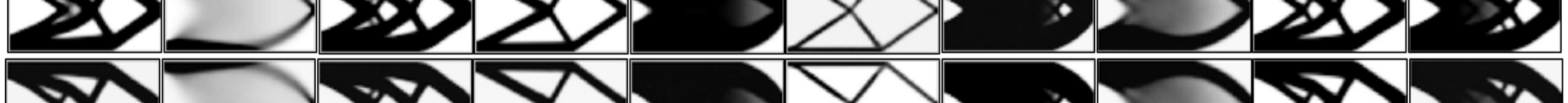 |
| Predicted images | 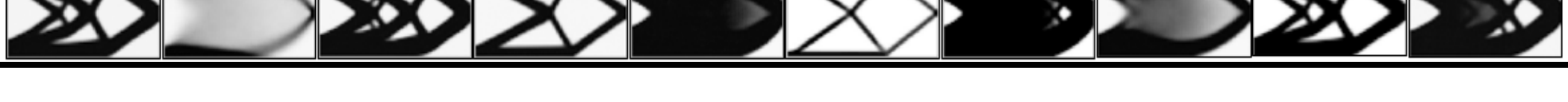 |

Then, the generated images by the WGAN whose size is 8×8 are transmitted to the decoder of CAE to complete generation task.

The inception scores for the proposed WGAN-CAE model and compared models are reported in Table 9 and representative images are compared in Fig. 13.

Table 9

Inception scores of WGAN-CAE, DCGAN, WGAN, and WGAN-GP.

| Metric | WGAN-CAE | DCGAN | WGAN | WGAN-GP |
|---|---|---|---|---|
| Inception score | 5.97 | 4.62 | 1.02 | 0.99 |

The proposed WGAN-CAE achieves the best inception scores. Compared with the DCGAN, the WGAN-CAE achieves 29.22% improvement in term of inception score.

As shown in Fig. 13, the images generated by the DCGAN lack convincing details and suffer blurred region in most cases, which makes them neither realistic enough nor have sufficiently high resolution. For WGAN and WGAN-GP, they are difficult to obtain the characteristics of the sample and their generated results seems helpless for contour image dataset.

Importantly, the WGAN-CAE does not achieve good results by original samples but by compressed samples. The characteristic of low dimensional data is well achieved, it can be inferred that the accuracy and quality of generated can be increased through dimensionality reduction.

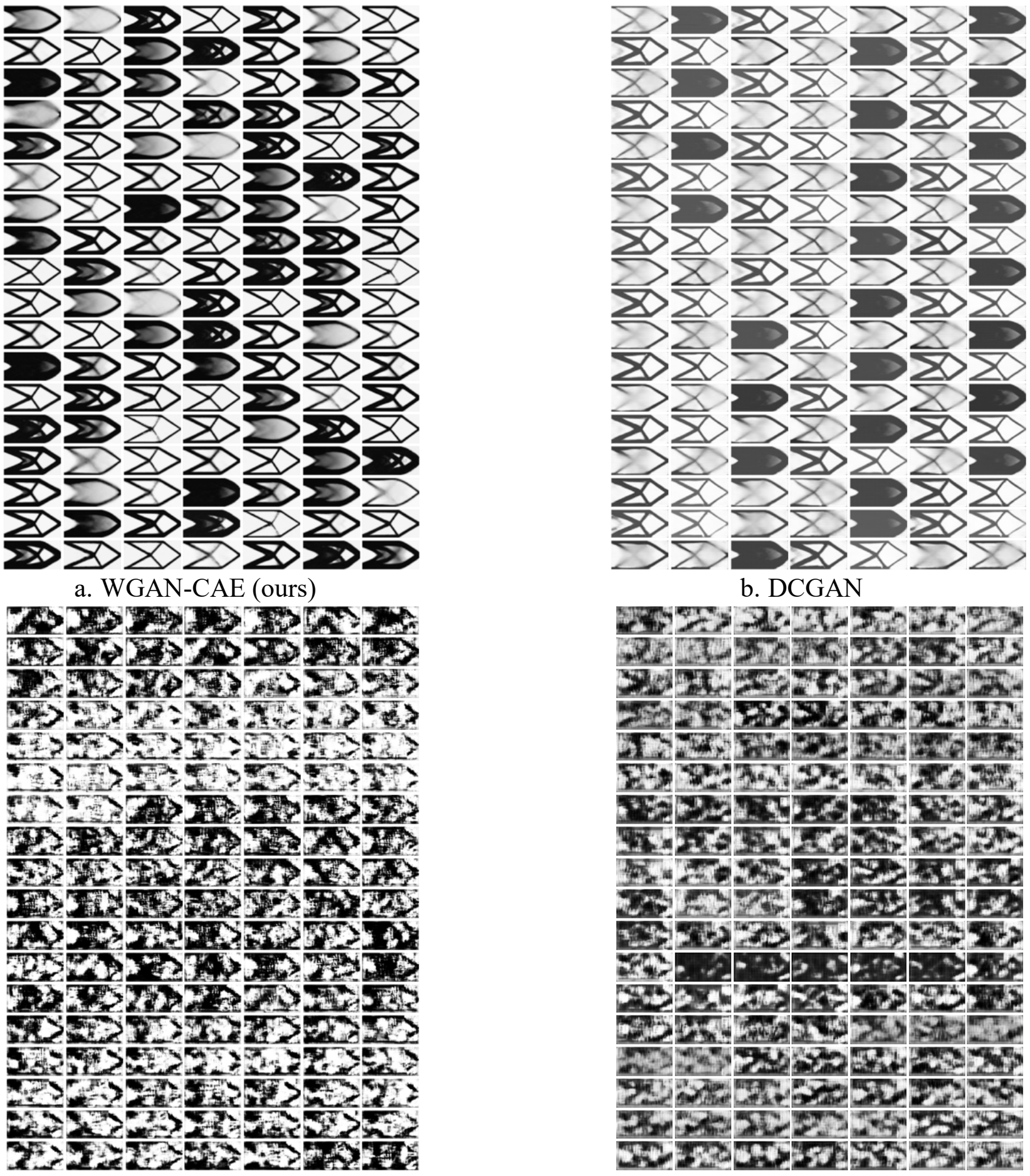

a. WGAN-CAE (ours) b. DCGAN

c. WGAN d. WGAN-GP

Fig. 13. Comparison of different GAN's results trained after 20 epochs.

The epoch is defined as the number of training steps after all samples have been trained once.

*4.2.3. Reconstruction of topology optimization process.*

Each compliance of pseudo image generated in Section 4.2.2 is calculated by trained CIC. Then the Lagrange polynomial expressed in Eqs. (39) and (40) is employed to complete the reconstruction task.

$$f(x)=\sum_{i=1}^{n} y_i p_i(x),\ i=1,2,\ldots,n \tag{39}$$

where

$$p_i = \prod_{i=1, i \neq j}^{n} \frac{x - x_i}{x_j - x_i} = \frac{(x - x_1)(x - x_2)\ldots(x - x_n)}{(x_j - x_1)\ldots(x_j - x_{j-1})(x_j - x_{j+1})\ldots(x_j - x_n)} \quad (40)$$

Finally, the reconstructed topology optimization process is shown in Fig. 14.

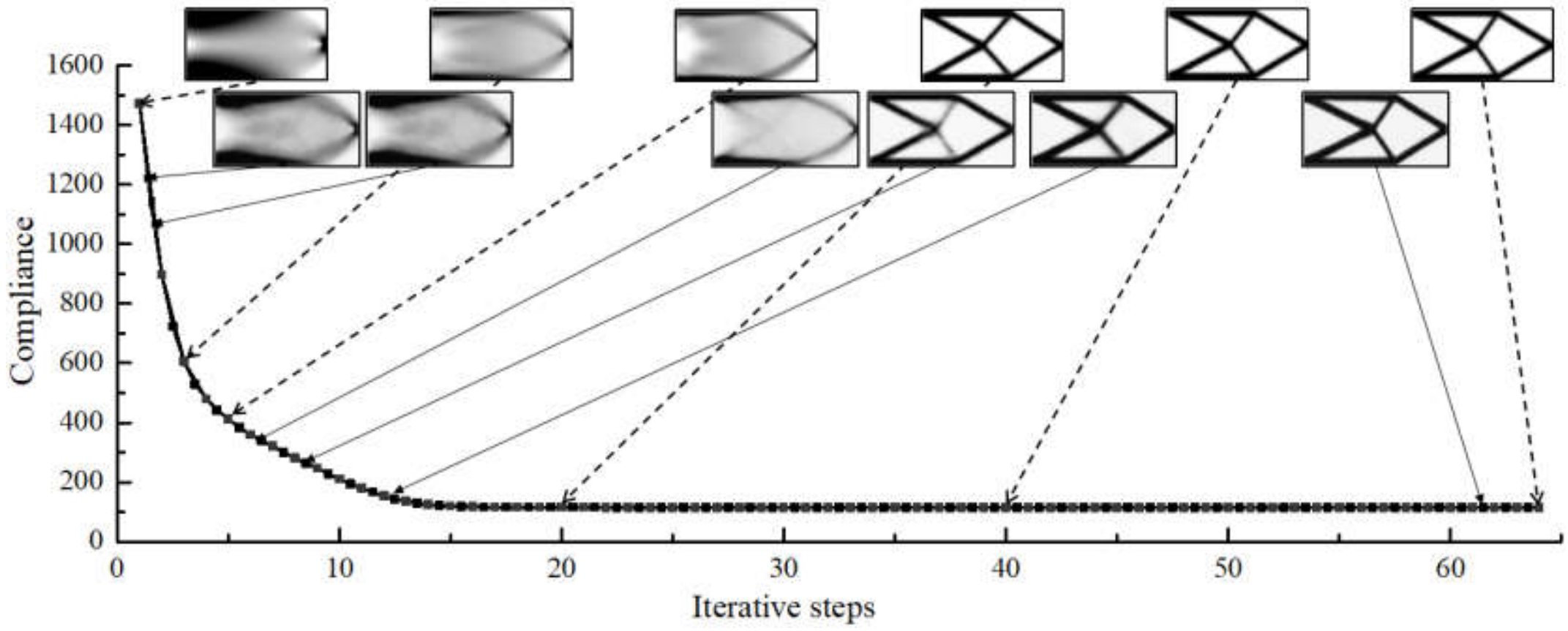

Fig. 14. The reconstruction of the topology optimization.

The images pointed using dotted arrow are the original topology optimization process, the total iteration steps is 64 and there are 64 corresponding contour images. The images pointed using solid arrow are the "pseudo contour images" generated by WGAN-CAE, and the number of contour images of reconstruction is extended to 6400.

## 5. Engineering Application

With the exponential increase in the power density of microelectronic components and their continuous miniaturization in overall dimensions, thermal management becomes a fundamental but pivotal element in electronic product design. How to efficiently visualize heat transfer process and remove the heat of the electronic equipment have been deemed as a major issue (91, 92). In this section, the ReConNN is applied to reconstruction of heat transfer process of a pin fin heat sink.

### *5.1. Finite element model*

The pin fin heat sink is a natural convection case with nine design variables. The purpose of the design is to minimize the highest temperature. The initial temperature of the whole system is set to 25℃ and the distribution of fits is centrosymmetric. As shown in Fig. 15, in order to simplified calculation, the pin fin heat sink is simplified from a 3-dimensional model to a 2-dimensional model along the A-A section.

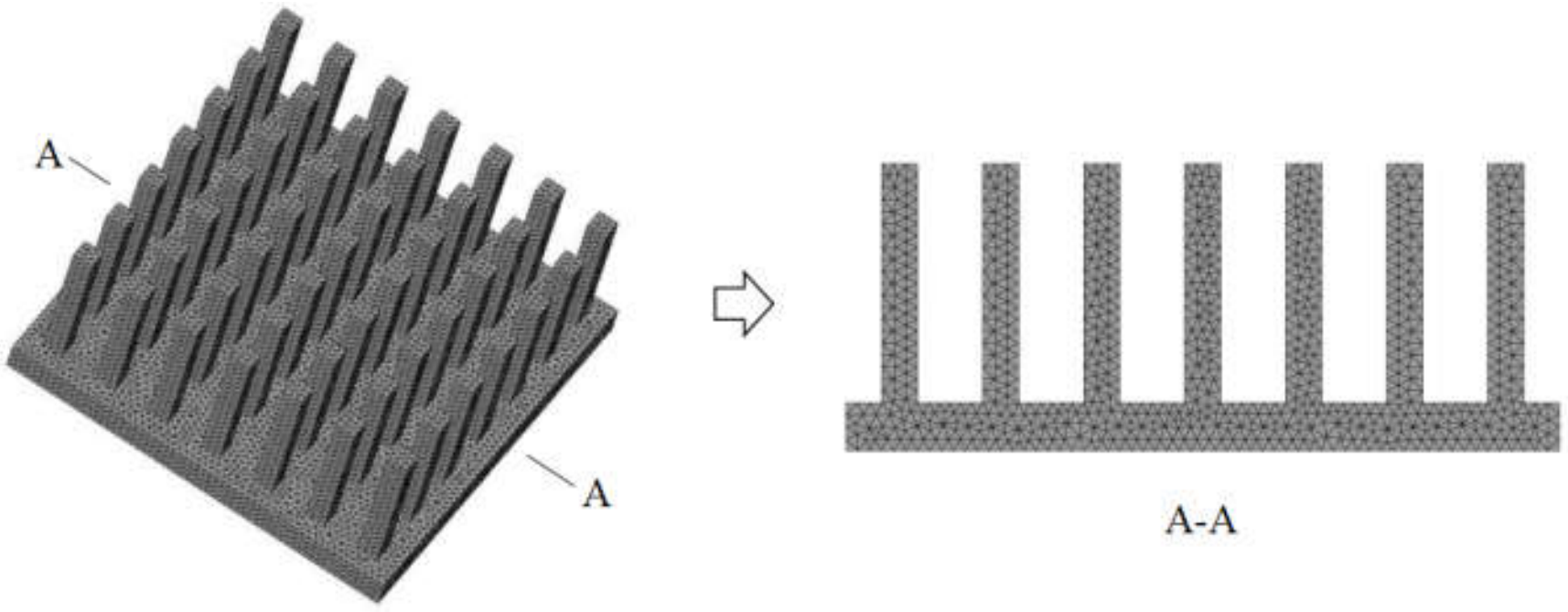

Fig. 15. Full-scale model of the pin fin heat sink.

The heat sink is made of aluminum alloy material. The fluid is air in the optimization. Effects of gravity and radiative heat transfer are neglected. The flow is steady and three dimensional. The governing equations of continuity, momentum, and energy in laminar flow are shown in Eqs. (41) - (43), respectively.

$$\nabla \cdot \vec{V} = 0 \tag{41}$$

$$\vec{V} \cdot \nabla \vec{V} = -\frac{1}{\rho}\nabla P + \nu \nabla^2 \vec{V} \tag{42}$$

$$\vec{V} \cdot \nabla T = \alpha \nabla^2 T \tag{43}$$

The energy equation for the solid parts is

$$\nabla^2 T = 0 \tag{44}$$

where $\vec{V}$ is the velocity vector representing the flow field, $\rho$ is the fluid density, $\nabla P$ is the convective pressure gradient, $v$ is the kinematic viscosity, $\alpha$ is the thermal diffusivity, and $T$ is the fluid temperature.

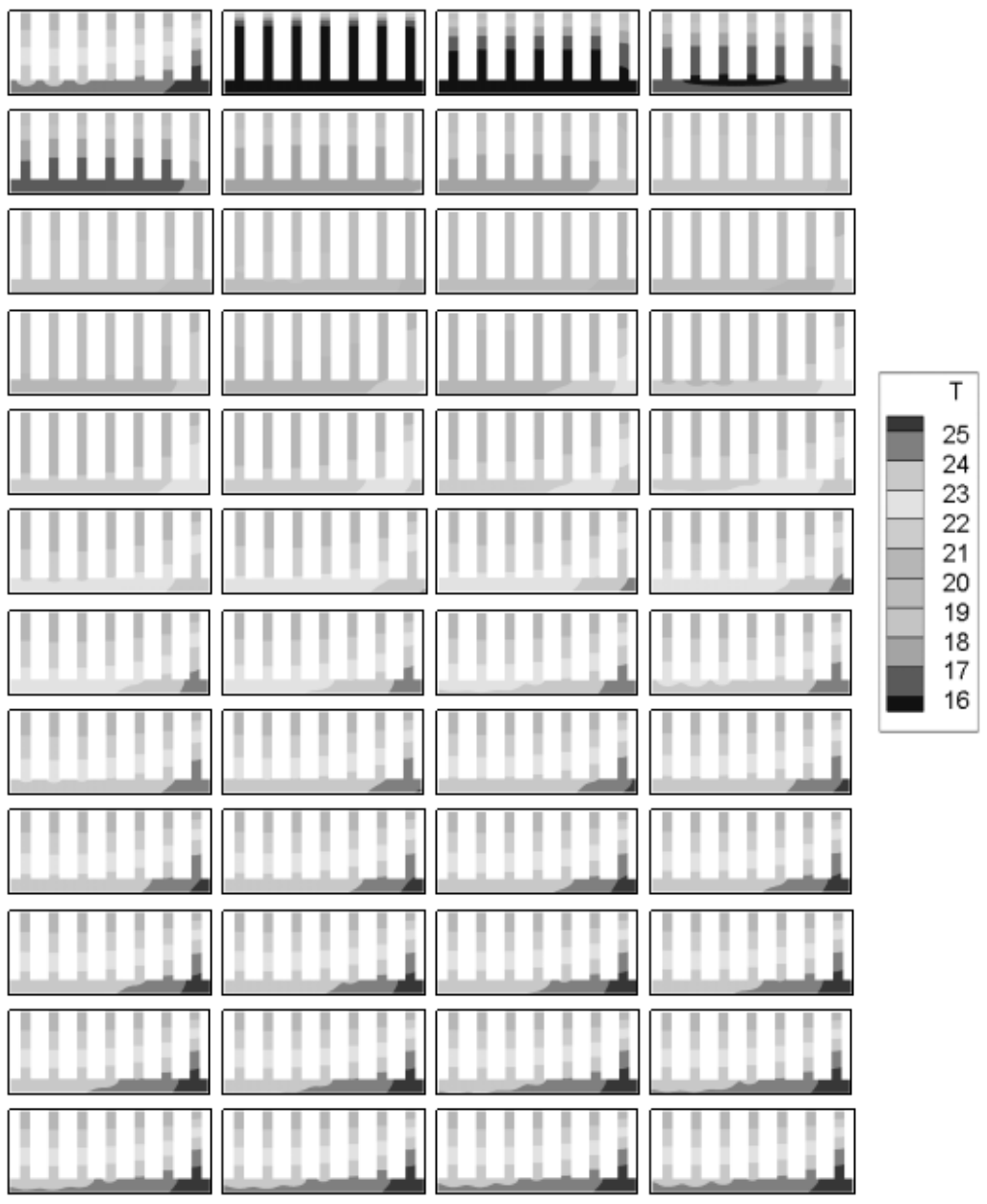

Fig. 16. Training samples of the pin fin heat sink.

*5.2. Reconstruction using ReConNN*

The number of training samples is 500 which is shown in Fig. 16. Training results of the CIC and generated images using WGAN-CAE are shown in Figs. 17 and 18 respectively. Finally, the reconstructed of time-based model of heat transfer process is shown in Fig. 19.

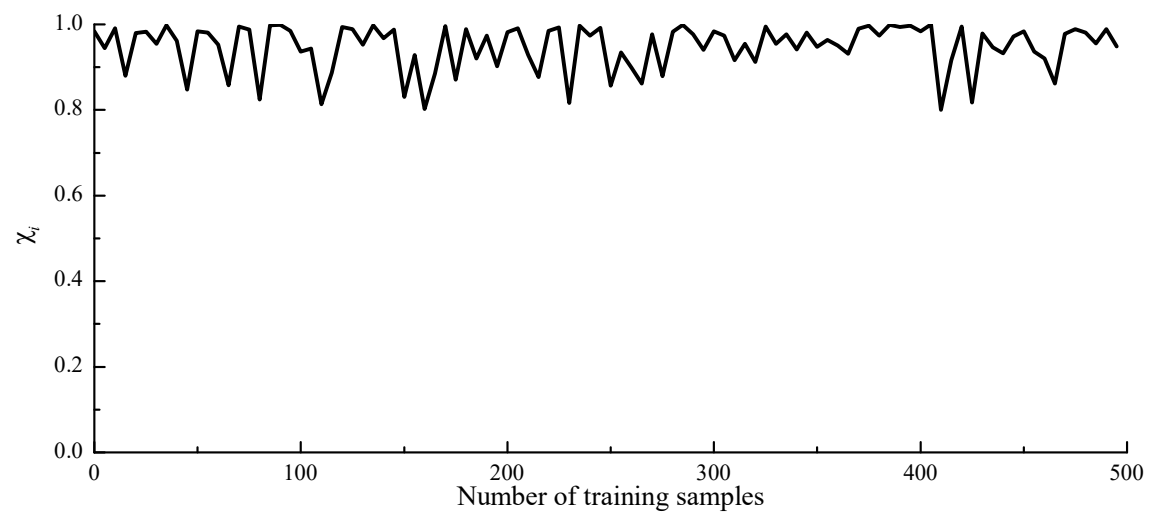

Fig. 17. Training results of CIC.

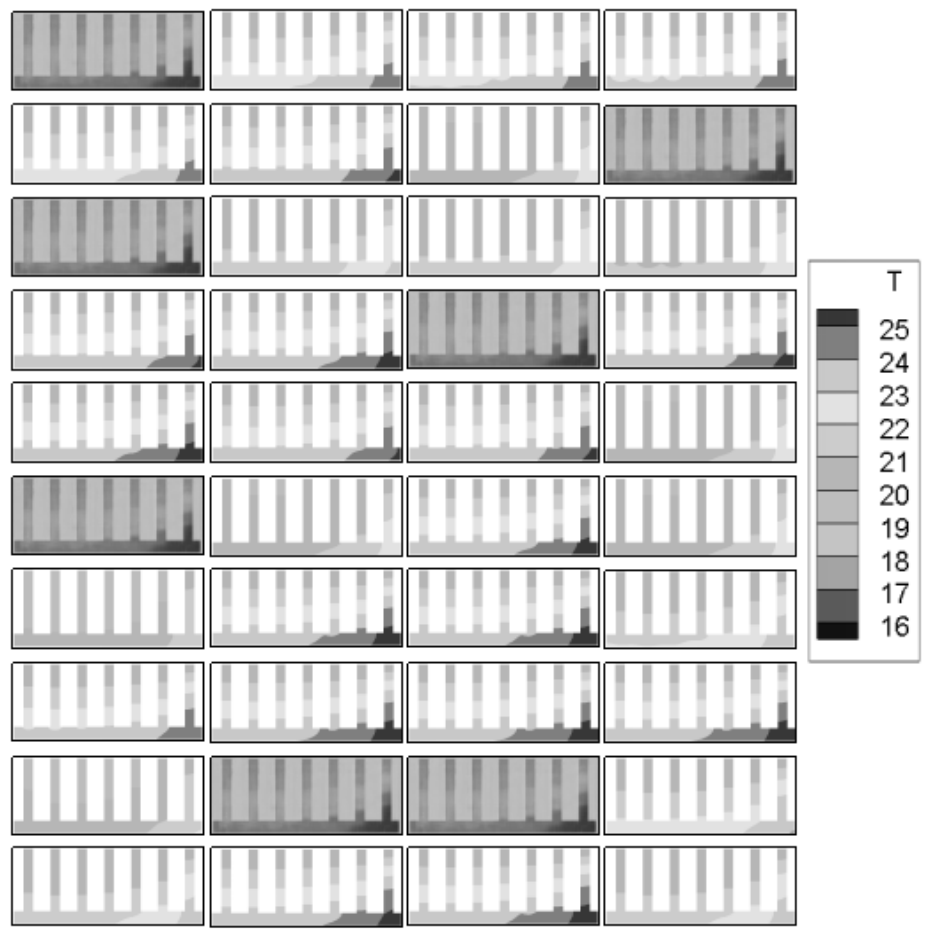

Fig. 18. Generated results using WGAN-CAE.

## 6. Conclusions and Future Work

In this study, a rReConNN model is developed for physical field reconstruction. This neural network proceeds from the framework "from image based model to analysis based model" and its advantages can be summarized as follows.

The CIC submodel is proposed with image incision technology for images regression tasks. Compared with existing CNN models, the CIC achieves a higher accuracy and a lower computational cost.

The WGAN-CAE submodel is proposed as a generative image model. Compared with other generative image models, the WGAN-CAE can generate higher accuracy images.

According to a classical topology optimization example, the results demonstrate that the proposed

ReConNN model has a potential capability to reconstruct model for the expensive evaluation problems.

Finally, the ReConNN model is applied to the reconstruction of heat transfer process of a pin fin heat sink. It is found that the proposed model can be applied to engineering applications well.

In addition, the proposed ReConNN model is not limited to the applications in this study. It can be further applied to multidisciplinary such as modeling for time serial problems or some experiments, such as to use DIC (Digital Image Correlation) camera to capture dynamic characteristics and reconstruct the model through these photos.

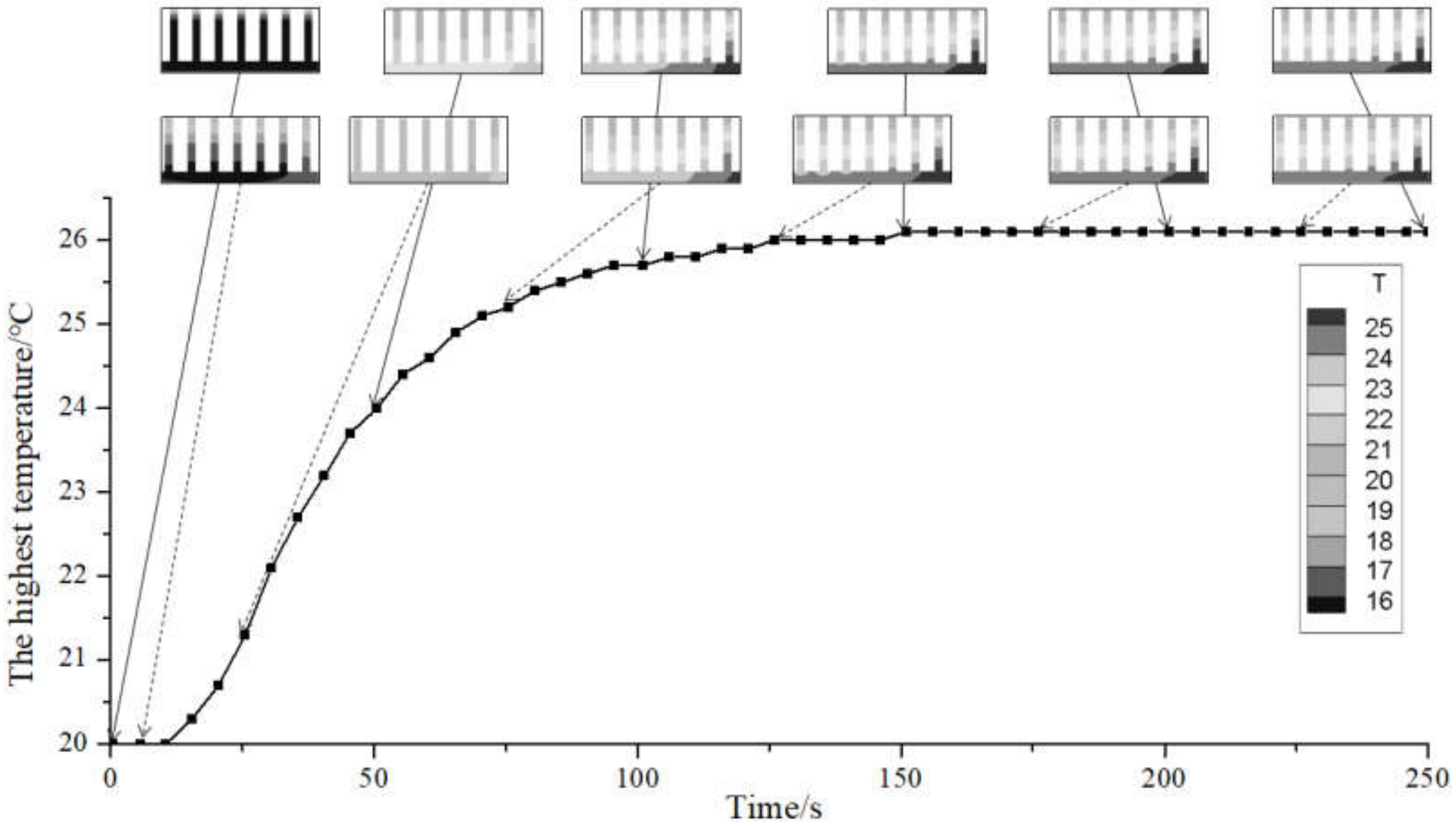

Fig. 19. The reconstruction of time-based model of the heat sink's heat transfer process.

The images pointed using solid arrow are the original heat transfer process, the total iteration steps is 500 and there are 500 corresponding contour images. The images pointed using dotted arrow are the "pseudo contour images" generated by WGAN-CAE, and the number of contour images of reconstruction is extended to 5000.

## Acknowledgments

This work has been supported by Project of the Key Program of National Natural Science Foundation of China under the Grant Numbers 11572120.